\documentclass{article} 
\usepackage{iclr2027_conference,times}

\usepackage{graphicx}
\usepackage{xspace}
\usepackage{xcolor}
\usepackage{amsmath}
\usepackage{amssymb}
\usepackage{booktabs}
\usepackage{caption}
\usepackage{subcaption}
\usepackage{enumitem}
\usepackage{microtype}
\usepackage{multirow}
\usepackage{tabularx}
\usepackage{nicefrac}
\usepackage{colortbl}
\usepackage{pifont}
\usepackage{placeins}
\usepackage{afterpage}
\usepackage{wrapfig}


\graphicspath{{figures/}}

\newcommand{\sysname}{\textsc{P3D-Bench}}

\newcommand{\Text}{Text-to-3D}
\newcommand{\Image}{Image-to-3D}
\newcommand{\TextImage}{Assembly-3D}
\newcommand{\TextDesc}{descriptive specification\xspace}
\newcommand{\TextParam}{parametric specification\xspace}
\newcommand{\TextDescShort}{Desc.\xspace}
\newcommand{\TextParamShort}{Param.\xspace}

\newcommand{\IoUV}{\ensuremath{\mathrm{IoU}_{\mathrm{V}}}}
\newcommand{\IoUC}{\ensuremath{\mathrm{IoU}_{\mathrm{C}}}}

\newcommand{\best}[1]{\textbf{\boldmath #1}}
\newcommand{\second}[1]{\underline{#1}}

\newcommand{\yes}{\textcolor{black!85}{\ding{51}}}
\newcommand{\no}{\textcolor{black!55}{\ding{55}}}
\newcommand{\nodash}{\textcolor{black!85}{\rule[0.4ex]{0.55em}{0.9pt}}}

\newcolumntype{Y}{>{\raggedright\arraybackslash}X}

\usepackage[hyphens]{url}
\usepackage{hyperref}
\hypersetup{
  breaklinks=true,hidelinks,
  pdftitle={P3D-Bench: Benchmarking MLLMs for Parametric 3D Generation and Structural Reasoning},
  pdfauthor={Yikang Yang; Zhanpeng Hu; Youtian Lin; Mengqi Zhou; Jingxi Xu; Feihu Zhang; Jiaheng Liu; Yao Yao}
}

\iclrfinalcopy

\title{\sysname{}: Benchmarking MLLMs for Parametric 3D Generation
       and Structural Reasoning}

\author{%
Yikang Yang\textsuperscript{1,$\dagger$}\quad
Zhanpeng Hu\textsuperscript{1,$\dagger$}\quad
Youtian Lin\textsuperscript{1}\quad
Mengqi Zhou\textsuperscript{1} \\[0.1em]
\bfseries Jingxi Xu\textsuperscript{2}\quad
Feihu Zhang\textsuperscript{2}\quad
Jiaheng Liu\textsuperscript{1}\quad
Yao Yao\textsuperscript{1,$\ddagger$} \\[0.1em]
{\mdseries\textsuperscript{1}Nanjing University \quad
\textsuperscript{2}Envision}%
}

\begin{document}

\vspace*{-28pt}
\maketitle
\fancyhead{}
\renewcommand{\headrulewidth}{0pt}
\fancyfoot[C]{\thepage}

\begingroup
\renewcommand{\thefootnote}{\fnsymbol{footnote}}
\footnotetext[2]{Equal contribution.\quad\textsuperscript{$\ddagger$}Corresponding author.}
\endgroup
\vspace{-12pt}

\begingroup
\setlength{\intextsep}{2pt}
\begin{figure}[!ht]
  \centering
  \includegraphics[width=0.90\linewidth,trim=18bp 0 23bp 0,clip]{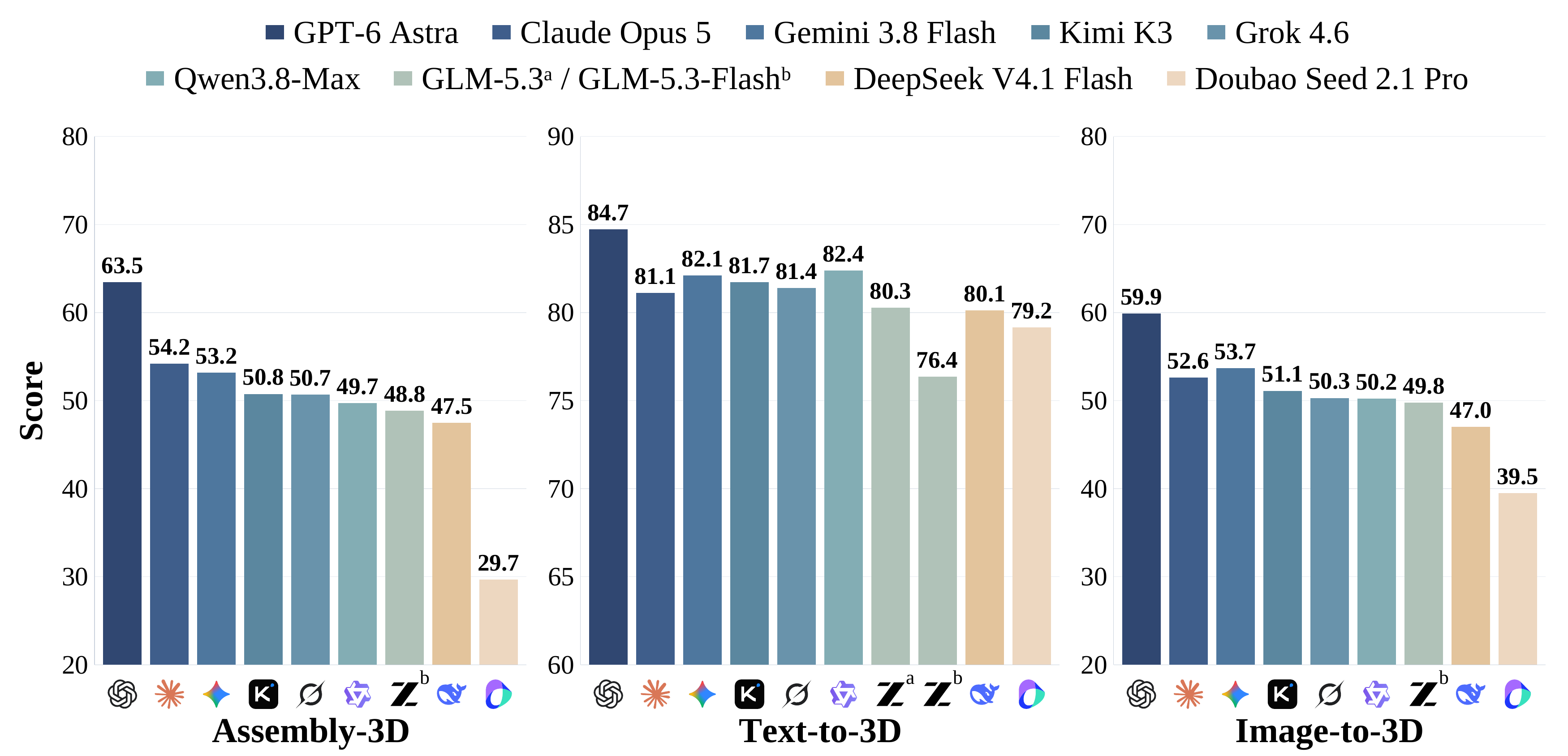}
  \caption{\textbf{Model scores across the three \sysname{} tasks.}
  Scores average the applicable Geo and Judge buckets, with Part additionally
  included for \TextImage{}.
  Per-bucket results are reported in Table~\ref{tab:main-results}.}
  \label{fig:teaser}
\end{figure}
\endgroup

\begin{abstract}
Multimodal large language models can write code to produce complex programs as well as use programs to do 3D modeling, which opens up a new avenue for 3D generation powered by their priors, world knowledge and reasoning.
Yet existing benchmarks rarely evaluate 3D modeling through code.
Such modeling demands more than runnable code: from a text or visual specification, a model must generate a parametric 3D program that is geometrically precise, semantically aligned and assembly-consistent.
We introduce \sysname{}, a benchmark for parametric 3D generation.
Unlike a 3D mesh, a parametric 3D program exposes explicit dimensions, construction operations and part relations, revealing whether a model recovers a design's structure, not just its appearance.
Under a unified protocol, \sysname{} covers three task families (Text-to-3D, Image-to-3D and Assembly-3D) and scores each output for executability, geometric fidelity, topology, text-grounded constraints, multiview semantic alignment and part-level structure.
We construct P3D-Dataset, comprising 400 text cases, 400 image cases, and 203 annotated assemblies.
Our evaluation on 100 cases from each task family yields three key findings.
First, multi-part generation is substantially more challenging than single-part modeling, with models struggling to compose individual parts into a coherent structure.
Second, models can often recover the global shape and semantic identity of the target object, yet fail to reproduce the precise parametric geometry specified by the input.
Third, part-level modeling remains weak on assemblies, where models recover neither the geometry of each part nor the right number of parts.
These results position \sysname{} as a benchmark for evaluating precise parametric geometry and part-level structure in parametric 3D generation.

\end{abstract}

\afterpage{%
  \begin{figure}[!t]
  \centering
  \includegraphics[width=\linewidth]{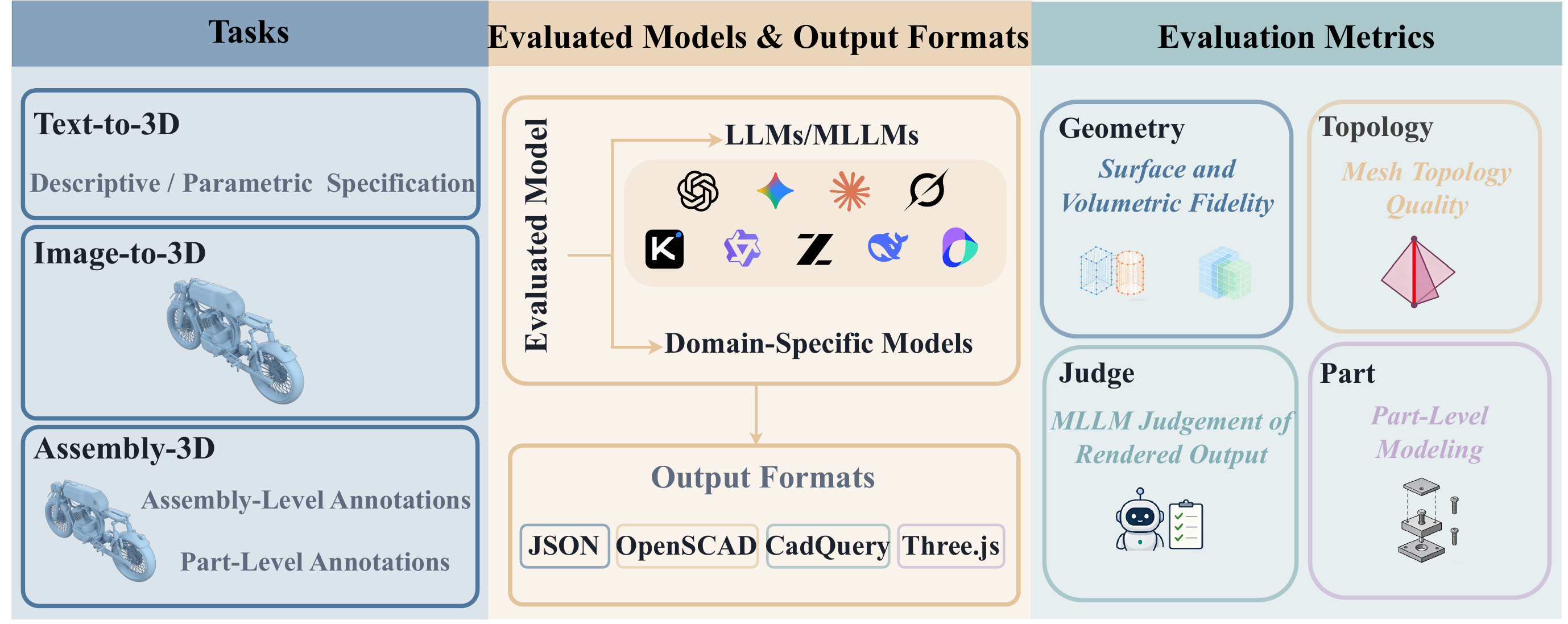}
  \caption{\textbf{Overview of the \sysname{} evaluation.} Given a task
  input---a text specification (descriptive or parametric), a single
  image or an image with assembly- and part-level annotations---a
  model (general-purpose LLM/MLLM or domain-specific) writes
  a program in one of the four output formats. \sysname{} executes and
  renders each program, reports validity (\emph{Valid}) and scores it
  along the four evaluation dimensions defined in the
  \emph{Evaluation protocol} section.}
  \label{fig:overview}
\end{figure}
  \afterpage{%
    \begin{table}[!t]
  \centering
  \caption{\textbf{Comparison of representative benchmarks.} Task describes the
  kind of task evaluated. Model Types indicates whether the benchmark evaluates
  general-purpose LLMs/MLLMs and domain-specific models (Domain Spec.).
  Under Evaluation, Exec. means that generated outputs are executed as 3D
  geometry; Param. means that scoring explicitly checks dimensional or
  parametric accuracy against the specification or ground truth; Spatial means that
  evaluation includes the relative placement or layout of objects or
  parts; Asm. means that the
  benchmark contains multi-part assemblies; and Part means explicit
  geometric matching of predicted and ground-truth parts.}
  \label{tab:benchmarks}
  \renewcommand{\arraystretch}{1.0}
  \setlength{\tabcolsep}{2pt}
  \footnotesize
  \begin{tabularx}{\linewidth}{@{}l>{\raggedright\arraybackslash}Xccccccc@{}}
    \toprule
    & & \multicolumn{2}{c}{Model Types} & \multicolumn{5}{c}{Evaluation} \\
    \cmidrule(lr){3-4}\cmidrule(lr){5-9}
    Benchmark & Task & \shortstack{LLM/\\MLLM} & \shortstack{Domain\\Spec.}
      & Exec. & Param. & Spatial & Asm. & Part \\
    \midrule
    SWE-bench & Code fix & \yes & \no & \no & \no & \no & \no & \no \\
    InfiniBench & Spatial reasoning & \yes & \no & \no & \no & \yes & \no & \no \\
    Theory of Space & Spatial reasoning & \yes & \no & \no & \no & \yes & \no & \no \\
    T\textsuperscript{3}Bench & Mesh gen. & \no & \yes & \no & \no & \no & \no & \no \\
    3DGen-Bench & Mesh gen. & \no & \yes & \no & \no & \no & \no & \no \\
    VoxelCodeBench & Procedural 3D gen. & \yes & \no & \yes & \no & \yes & \no & \no \\
    3DCodeBench & Procedural 3D gen. & \yes & \no & \yes & \no & \no & \no & \no \\
    Text2CAD-Bench & CAD gen. (CadQuery) & \yes & \yes & \yes & \no & \no & \no & \no \\
    UniCAD & CAD gen. (CadQuery) & \yes & \yes & \yes & \no & \yes & \no & \no \\
    BenchCAD & CAD gen. (CadQuery) & \yes & \yes & \yes & \yes & \no & \no & \no \\
    MUSE & CAD gen. (CadQuery) & \yes & \no & \yes & \no & \yes & \yes & \no \\
    \midrule
    \textbf{\sysname{} (Ours)} & \textbf{CAD gen. (JSON, OpenSCAD, CadQuery, Three.js)}
      & \yes & \yes & \yes & \yes & \yes & \yes & \yes \\
    \bottomrule
  \end{tabularx}
\end{table}
    \afterpage{
      \afterpage{%
\begin{figure}[!t]
  \centering
  \includegraphics[width=0.98\linewidth]{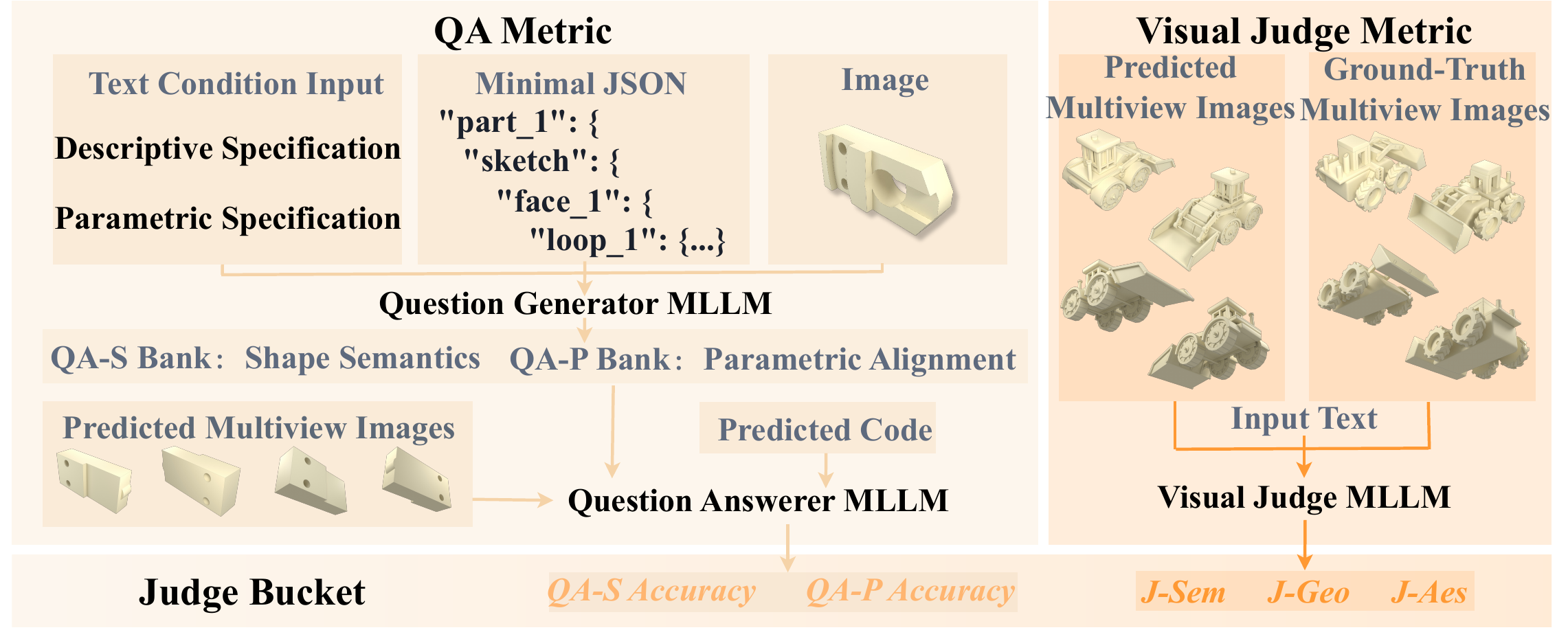}
  \caption{\textbf{Overview of the \emph{Judge} bucket metrics.}
  The QA metric builds a QA-S bank from the descriptive specification and
  a QA-P bank from the parametric specification, then scores the prediction
  by its accuracy on those questions. The visual Judge rates the prediction
  from the predicted and ground-truth multiview images and, when available,
  the input text, along three axes: semantic similarity (\emph{J-Sem}),
  geometric similarity (\emph{J-Geo}) and aesthetic quality (\emph{J-Aes}).}
  \label{fig:metric-judge}
\end{figure}
        \afterpage{
\begin{figure}[!t]
  \centering
  \includegraphics[width=0.98\linewidth]{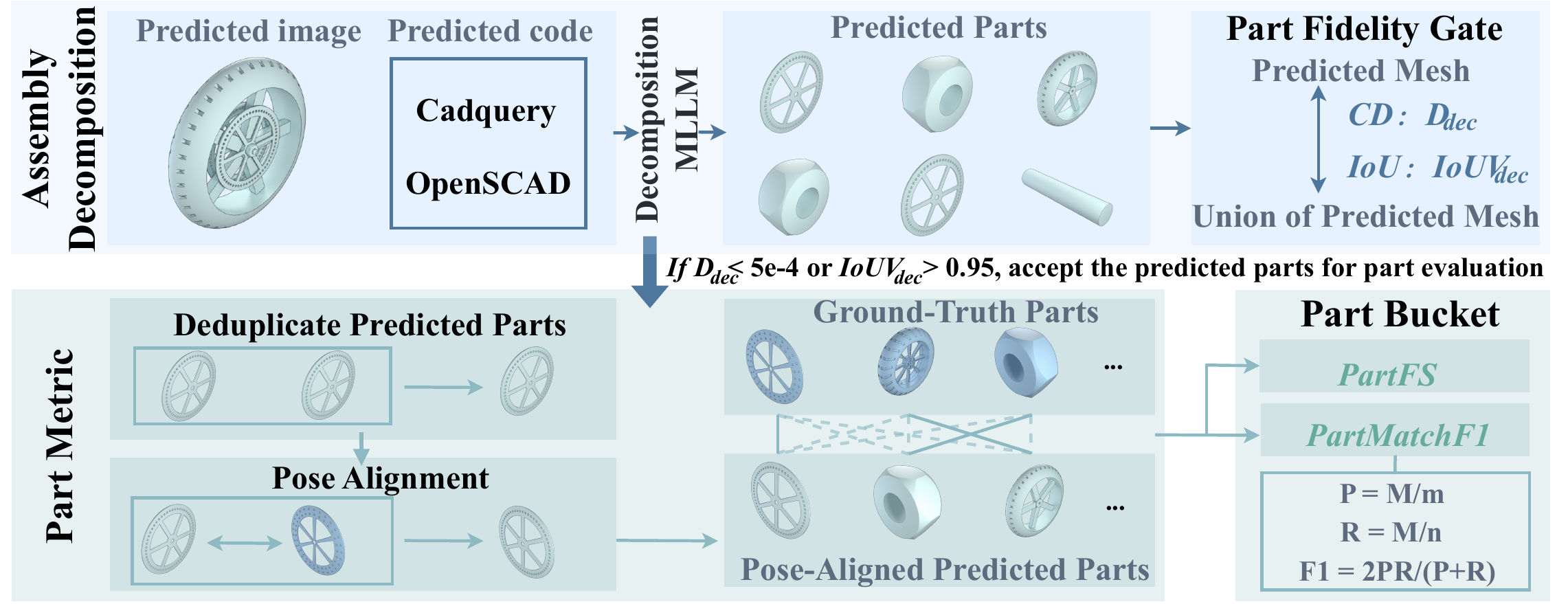}
  \caption{\textbf{Overview of the Assembly-3D \emph{Part} bucket metrics.}
  \emph{Assembly Decomposition}: a fixed decomposition MLLM splits the
  predicted assembly into parts, and a fidelity gate keeps only those whose
  reassembled union matches the original prediction. \emph{Part Metric}:
  the retained parts are deduplicated, pose-aligned to GT parts and matched
  one-to-one to compute the scores, where \(M\) counts the successful
  matches and \(m\), \(n\) count the predicted and GT parts.}
  \label{fig:metric-part}
\end{figure}
}%
      }%
    }%
  }%
}

\section{Introduction}
\label{sec:introduction}

Multimodal large language models (MLLMs) can write executable code and reason about the shapes and structures in images.
Combining these abilities, current models can turn a text prompt or reference image into CadQuery or Three.js code that executes into the target 3D object.
Unlike directly predicting a mesh, generating 3D models as programs gives an explicit, editable representation.
The code records the dimensions, construction steps and decomposition into parts, and can be rerun with different parameters to produce a new valid model.

How well current models perform at parametric 3D generation remains an open question.
Existing benchmarks emphasize complementary capabilities:
code benchmarks check whether a program compiles and passes its tests~\citep{chen2021codex,jimenez2024swebench};
spatial benchmarks assess reasoning about layout and object relations~\citep{wang2025infinibench,zhang2026theoryofspace};
text-to-3D benchmarks assess the shape's visual quality~\citep{he2023t3bench,zhang2025threedgenbench}.
Parametric 3D generation requires these abilities jointly: the generated program must compile into valid geometry and, by inferring part structure and assembly relations, produce a result that is both visually plausible and geometrically precise.

We introduce \sysname{}, a benchmark that evaluates these properties of 3D modeling through code across three tasks.
\textbf{\Text{}} provides a text description and requires a single part.
\textbf{\Image{}} provides a single rendered image and requires a multi-part object, so the model must infer geometry and part layout not visible from one view.
\textbf{\TextImage{}} adds assembly-level and part-level text annotations and requires the full assembly.
In each task, the model produces a program in one of four formats (JSON~\citep{khan2024text2cad}, OpenSCAD~\citep{openscad}, CadQuery~\citep{cadquery} or Three.js~\citep{threejs}), which \sysname{} executes and renders; it reports executable validity and scores the model from four perspectives: geometry, topology, MLLM-based judgment and part-level structure (Figure~\ref{fig:overview}).

Existing datasets are uneven in quality and lack the structured annotations these tasks require, so we build P3D-Dataset.
Starting from Text2CAD~v1.1~\citep{khan2024text2cad} and the Fusion~360 Gallery~\citep{willis2022joinable}, we filter ambiguous and near-duplicate models, balance complexity levels, and generate and validate the task-specific text inputs: descriptive and parametric specifications for \Text{}, and assembly-level and part-level annotations for \TextImage{}.
The resulting dataset comprises 400 text cases, 400 image cases and 203 annotated assemblies.

In summary, \sysname{} provides a unified benchmark for parametric 3D generation and structural reasoning from text and image specifications.
Extensive evaluations of frontier MLLMs, text-only LLMs and domain-specific models reveal that:
(1) building a full assembly is substantially harder than building a single part because it requires correct spatial relations and part structure.
(2) geometric alignment with the input is harder than semantic alignment: the strongest MLLM already aligns well with the input semantically (\emph{J-Sem}\,\(\approx\)\,0.7), but its geometry matches the specified dimensions less accurately (\emph{J-Geo}\,\(\approx\)\,0.5).
(3) on assembly tasks, beyond errors in part placement, models also struggle to recover the correct number and geometry of parts: matching predicted parts against ground-truth parts, the strongest MLLM reaches a \emph{PartMatchF1} of about \(0.5\).
We make four contributions.

\begin{itemize}[leftmargin=*, topsep=2pt, itemsep=1pt, parsep=0pt, partopsep=0pt]
  \item \textbf{A unified benchmark.}
  Three tasks (\Text{}, \Image{} and \TextImage{}) and four code formats (JSON, OpenSCAD, CadQuery, Three.js) under a unified protocol that executes and renders every output, enabling comparison across input specifications and target formats.
  \item \textbf{A new dataset and its pipeline.}
  A filtering and annotation pipeline yielding P3D-Dataset: 400 text cases, 400 image cases and 203 assemblies with task-specific text inputs.
  \item \textbf{A structured evaluation protocol.}
  Beyond validity, four scores (\emph{Geo}, \emph{Topo}, \emph{Judge}, \emph{Part}) covering geometric fidelity, topology, MLLM-based assessment and part structure.
  \item \textbf{An evaluation of frontier models.}
  A unified evaluation of frontier MLLMs, text-only LLMs and domain-specific models, revealing that programs which execute and render plausibly still fail to recover correct parametric geometry.
\end{itemize}

\section{Related work}
\label{sec:related_work}

\paragraph{Visual and parametric 3D generation.}
Most 3D generation produces visual geometry.
DreamFusion~\citep{poole2023dreamfusion} and LRM~\citep{hong2024lrm} build on the NeRF representation~\citep{mildenhall2020nerf} to recover a radiance field from text or a single image, while native 3D diffusion transformers~\citep{wu2024direct3d,wu2025direct3ds2} build on neural signed distance fields~\citep{park2019deepsdf} to generate meshes directly.
Unlike these appearance-only outputs, parametric 3D generation targets a representation defined by its construction logic, making operations, dimensions and part structure explicit and editable.
Early models are trained to generate sketch-extrude sequences directly~\citep{wu2021deepcad,xu2022skexgen,xu2023hierarchical,khan2024text2cad}.
Later work fine-tunes code LLMs and VLMs to produce executable CAD code~\citep{doris2025cadcoderopensourcevisionlanguagemodel,xie2025texttocadquery,govindarajan2026cadmium,kolodiazhnyi2026cadrille}.
Recent agentic systems clarify ambiguous prompts before generation~\citep{clarify2026} or revise CAD code under geometric and visual feedback~\citep{cadsmith2026}; ArtiCAD and ArtiCraft extend agentic generation to articulated assemblies with explicit joints~\citep{shui2026articad,zhou2026articraft}.
A parallel line of work generates Blender Python instead of CAD code, with agents that write and iteratively repair the script from text~\citep{lu2025ll3m} or a single image~\citep{yin2026vision} under rendered feedback.
Across these threads, parametric 3D is increasingly produced by general-purpose models writing code, motivating a unified evaluation of general-purpose LLMs/MLLMs and domain-specific models across tasks and code formats.

\paragraph{Benchmarks related to parametric 3D generation.}
Recent benchmarks for code-based and parametric 3D generation differ in their target representations, supported tasks and evaluation criteria.
Code-based 3D benchmarks operate on generated programs: VoxelCodeBench~\citep{zheng2026voxelcodebench} evaluates code against an Unreal voxel API, and 3DCodeBench~\citep{gao2026threedcodebench} scores agentic Blender Python by executability, render similarity, human preference and geometric similarity.
Both verify that the 3D program runs, but neither reports dedicated metrics for numerical dimensional accuracy or geometric part matching.
Another line evaluates executable CAD code directly.
Text2CAD-Bench~\citep{wang2026text2cadbench} focuses on text-to-CadQuery generation, while UniCAD~\citep{chen2026unicadunifiedbenchmarkuniversal} covers multiple input modalities and BenchCAD~\citep{zhang2026benchcad} also evaluates editing tasks.
Text2CAD-Bench measures validity, Chamfer distance and IoU on its L1--L3 levels and uses a VLM to assess multiview renders on L4; BenchCAD also tests numerical CAD questions.
These benchmarks use CadQuery and focus on single-body models rather than explicit geometric matching of assembly parts.
MUSE~\citep{dong2026muse} is the most related: it checks whether generated text-to-CadQuery assemblies are functional and assemblable. Its assembly structure, however, is judged by a VLM from rendered images.
Table~\ref{tab:benchmarks} summarizes these differences: unlike prior benchmarks, \sysname{} evaluates each case across multiple formats and includes multi-part assemblies with explicit part-level modeling.

\section{\sysname{}: Tasks, dataset and evaluation}
\label{sec:dataset}

\subsection{Task definition}
\label{sec:tasks}
We formulate parametric 3D generation as code-based reconstruction from a task condition \(c\)---a text description, one rendered image or the image together with assembly-level and part-level text.
Given an input \(c\) and a target format \(\phi\), a policy \(\pi\) generates a program \(f_\pi=\pi(c,\phi)\), where \(\phi\) is one of the four formats supported by \sysname{}: minimal JSON, OpenSCAD, CadQuery or Three.js.
A format-specific deterministic execution operator \(\mathcal{E}_{\phi}\) then compiles, executes and renders the resulting program to produce a 3D output \(y_\pi=\mathcal{E}_{\phi}(f_\pi)\).
Accordingly, we represent each evaluation case as a triplet \(x=(c,f^{\star},y^{\star})\), where \(f^{\star}\) is the source program and \(y^{\star}\) is its target 3D output, including ground-truth geometry and, when available, part structure.




\subsection{Dataset construction pipeline}
\label{sec:data_pipeline}

To evaluate the \sysname{} tasks, we construct P3D-Dataset from two open CAD sources with complementary geometry: Text2CAD~v1.1~\citep{khan2024text2cad}, which contains \(176{,}017\) single-part sketch--extrude programs for \Text{}, and the Fusion~360~Gallery~\citep{willis2022joinable}, which contains \(8{,}251\) multi-part assemblies for \Image{} and \TextImage{}.
Neither source is directly usable.
Text2CAD includes unevaluable, simple and near-duplicate records, and does not distinguish descriptive from parametric text specifications.
Fusion~360 needs screening for visual clarity and reconstructability; it also lacks the assembly-level and part-level text required by \TextImage{}.
We therefore design a two-stage pipeline.
The filtering stage removes unreliable and redundant cases and balances the retained set across category and complexity.
The annotation stage supplies the task-specific text inputs and keeps them consistent with the source geometry and render.
Appendix~\ref{app:dataset_details} summarizes the process and final splits.

\paragraph{Filtering Pipeline.}
Filtering constructs a balanced subset where failures reflect model limitations rather than broken sources, ambiguous inputs or repeated shapes.
We retain cases that are executable, renderable, visually interpretable and non-redundant, while balancing complexity across the dataset.
Specifically, deterministic checks first remove unevaluable records;
a review MLLM then rejects cases that are visually ambiguous, not reconstructable from the task input or poorly oriented, while assigning category and complexity labels; and
DINOv2~\citep{oquab2024dinov2} feature matching over render embeddings removes near-duplicate cases.
Finally, complexity-balanced sampling draws the retained cases across semantic categories and easy, medium and hard tiers, yielding \(400\) \Text{} cases and \(400\) \Image{} cases.


\paragraph{Annotation Pipeline.}
Annotation converts retained geometry into the task-specific inputs absent from the raw sources.
\((1)\) \emph{Text-to-3D annotation.}
For each Text2CAD part, an annotation MLLM writes two specifications from the source geometry and render.
The descriptive specification summarizes shape, salient features and function without numerical parameters, testing semantic shape construction from natural language.
The parametric specification expresses dimensions, counts, offsets and placements in natural language, testing whether models can translate explicit parameters into executable geometry.
\((2)\) \emph{Image-to-3D annotation.}
For each Fusion~360 assembly, the annotation MLLM labels each part after geometric deduplication (assemblies with more than \(20\) unique parts are dropped), then combines these part-level annotations with the assembly's spatial layout to compose an assembly-level description; a verification MLLM checks both for consistency and render alignment.
Only verified assemblies are retained for \TextImage{}, yielding \(203\) annotated cases.

\subsection{Evaluation protocol}
\label{sec:method}
\label{sec:metrics}



Executable parametric 3D generation can fail at several levels: a program may not execute, the resulting shape may be geometrically inaccurate, the generated mesh may have poor topology, the render may miss semantic or parametric constraints, and an assembly may use the wrong parts despite a plausible overall shape.
\sysname{} therefore executes each generated program and exports the result to a mesh.
Before scoring, we apply task-specific normalization to the predicted and ground-truth meshes, then align the prediction to the ground truth.

We measure executable validity (\emph{Valid}) and group the quality metrics into \emph{Geometry}, \emph{Topology}, \emph{Judge} and, for \TextImage{}, \emph{Part}.
\label{sec:buckets}
Each group combines its applicable sub-metrics into a \emph{bucket score} normalized to $0$--$100$, where higher is better.
The preprocessing and score aggregation procedures are detailed in Appendix~\ref{app:evaluation_details}.

\paragraph{Geometry and Topology Metrics.}
\label{sec:metrics-geom-topo}



The \emph{Geometry} metrics measure how closely an executable prediction matches the ground truth after alignment, using complementary surface and volume criteria: Chamfer Distance (CD$\downarrow$), F-score at the fine threshold (F@$0.01$$\uparrow$), normal consistency (NC$\uparrow$) and IoU (computed per task; see Appendix~\ref{app:mesh_alignment_details}).
To measure topological quality of meshes, we report the no-open-edge score (NoOE$\uparrow$), inverted-normal ratio (InvN$\downarrow$) and non-manifold-edge ratio (NM$\downarrow$).

\paragraph{MLLM Judge Metrics.}
\label{sec:metrics-mllm}


Since geometry and topology metrics do not capture semantic and parametric correctness, we introduce an MLLM-based \emph{Judge} bucket (Gemini~3.8~Flash~\citep{google2026gemini38flash}) with two components (Figure~\ref{fig:metric-judge}).
The QA metric tests the prediction using questions derived from the textual specification: QA-S targets semantic constraints in the descriptive setting, while QA-P targets explicit dimensions, counts, placements and other parametric constraints in the parametric setting.
The Visual Judge metric compares predicted and ground-truth multiview images along three dimensions: \emph{J-Sem}, \emph{J-Geo} and \emph{J-Aes}; the input text specification, when available, is also given to the judge.
\emph{J-Sem} evaluates object identity and semantic consistency, \emph{J-Geo} compares shape, proportions and alignment, and \emph{J-Aes} rates visual quality and detail richness.
We validate both components against human annotators on the same QA-S items and three axes (Table~\ref{tab:judge-human}).


\paragraph{Assembly-3D Part Metrics.}
\label{sec:metrics-part}

The \emph{Part} metrics evaluate per-part modeling ability: whether the predicted parts match the ground-truth parts in both shape and count.
Since evaluated models emit a single whole-assembly program, we use a fixed decomposition MLLM, shared across all models, to split each valid predicted assembly into per-part programs (Figure~\ref{fig:metric-part}).
A fidelity gate then checks that the decomposed parts reassemble to the original prediction before any per-part scores are computed; cases that fail the gate are excluded rather than scored.
Appendix~\ref{app:textimage_supplements} reports per-model decomposition fidelity and gate-exclusion counts.

After the fidelity gate, we deduplicate the retained predicted parts, align their poses to the ground-truth parts and match them using Hungarian matching~\citep{kuhn1955hungarian} on the pairwise part F-scores.
Let \(\mathcal{H}\) denote the set of one-to-one pairs returned by Hungarian matching.
For a prediction with \(m\) distinct parts and a ground-truth assembly with \(n\) distinct parts, \(|\mathcal{H}|=\min(m,n)\).
We then compute two part-level scores, PartFS and PartMatchF1:
\begin{equation}
  \mathrm{PartFS}=\frac{1}{\min(m,n)}\sum_{(p,g)\in\mathcal{H}} F_{\mathrm{part}}(p,g),
  \qquad
  \mathrm{PartMatchF1}=\frac{2\,P\,R}{P+R}.
  \label{eq:part}
\end{equation}
PartFS is the mean part F-score over all pairs in \(\mathcal{H}\), measuring the geometric fidelity of matched parts.
A pair is considered a \emph{successful match} if its part F-score is at least the acceptance threshold.
Let \(M\) denote the number of such successful matches.
PartMatchF1 is then the F1 score of part precision \(P=M/m\) (PartMatchP) and part recall \(R=M/n\) (PartMatchR).

\FloatBarrier

\begin{table}[!t]
  \centering
  \caption{\textbf{\sysname{} main results, with bucket scores averaged
  across output formats.}
  \emph{Geo}, \emph{Topo}, \emph{Judge} and \emph{Part} measure geometry,
  topology, MLLM judgment and assembly part quality;
  \emph{Valid} is executable validity (\%).
  Each model is evaluated only on its applicable tasks;
  ``\nodash'' denotes an unevaluated task--model pair.
  Best and second-best general-purpose models are bold and underlined,
  respectively, for \emph{Geo}, \emph{Judge} and \emph{Part}.}
  \label{tab:main-results}
  \setlength{\tabcolsep}{1.2pt}
  \renewcommand{\arraystretch}{0.95}
  \small
  \newcommand{\mainhead}[1]{{\fontsize{8}{9}\selectfont #1}}
  \newcommand{\mainvalid}{\mainhead{Valid}}
  \begin{tabular}{@{}>{\fontsize{8}{9}\selectfont}l@{\hspace{4pt}}
    >{\columncolor{blue!6}}c>{\columncolor{blue!6}}c>{\columncolor{blue!6}}ccc@{\hspace{4pt}}
    >{\columncolor{blue!6}}cc@{\hspace{4pt}}>{\columncolor{blue!6}}c>{\columncolor{blue!6}}ccc@{\hspace{4pt}}
    >{\columncolor{blue!6}}c>{\columncolor{blue!6}}ccc@{}}
    \toprule
    & \multicolumn{5}{c}{\mainhead{\textbf{\TextImage{}}}} & \multicolumn{6}{c}{\mainhead{\textbf{\Text{}}}} & \multicolumn{4}{c}{\mainhead{\textbf{\Image{}}}} \\
    \cmidrule(lr){2-6}\cmidrule(lr){7-12}\cmidrule(lr){13-16}
    & \multicolumn{5}{c}{} & \multicolumn{2}{c}{\mainhead{Descriptive}} & \multicolumn{4}{c}{\mainhead{Parametric}} & \multicolumn{4}{c}{} \\
    \cmidrule(lr){7-8}\cmidrule(lr){9-12}
    Model & \mainhead{Geo} & \mainhead{Judge} & \mainhead{Part} & \mainhead{Topo} & \mainvalid & \mainhead{Judge} & \mainvalid & \mainhead{Geo} & \mainhead{Judge} & \mainhead{Topo} & \mainvalid & \mainhead{Geo} & \mainhead{Judge} & \mainhead{Topo} & \mainvalid \\
    \midrule
      GPT-6~Astra & \best{68.2} & \best{59.9} & \best{62.3} & 95.8 & 99.4\% & \best{92.6} & 100\% & \second{68.8} & \best{92.8} & 99.3 & 100\% & \best{58.9} & \best{60.9} & 92.2 & 99.6\% \\
      Claude~Opus~5 & \second{56.3} & 52.0 & \second{54.3} & 98.5 & 100\% & 84.6 & 100\% & 67.6 & 91.2 & 99.5 & 100\% & 52.1 & 53.1 & 92.4 & 100\% \\
      Gemini~3.8~Flash & 56.1 & \second{52.9} & 50.5 & 97.4 & 99\% & 86.6 & 100\% & \best{69.2} & 90.5 & 99.3 & 100\% & \second{53.1} & \second{54.3} & 89.0 & 99\% \\
      Grok~4.6 & 55.0 & 49.6 & 47.5 & 97.6 & 100\% & 85.9 & 100\% & 68.7 & 89.6 & 99.5 & 100\% & 50.0 & 50.6 & 90.5 & 99\% \\
      Kimi~K3 & 52.8 & 48.3 & 51.2 & 94.9 & 99\% & 86.9 & 98.5\% & 67.0 & \second{91.3} & 98.8 & 99.5\% & 50.7 & 51.5 & 92.5 & 100\% \\
      Qwen3.8-Max & 52.3 & 47.3 & 49.6 & 95.9 & 98.9\% & \second{88.6} & 100\% & 67.9 & 90.7 & 99.0 & 100\% & 50.3 & 50.2 & 94.4 & 99.7\% \\
      GLM-5.3 & \nodash & \nodash & \nodash & \nodash & \nodash & 85.2 & 99.5\% & 65.6 & 90.0 & 99.5 & 100\% & \nodash & \nodash & \nodash & \nodash \\
      GLM-5.3-Flash & 52.6 & 48.1 & 45.9 & 96.2 & 98.9\% & 81.1 & 97.5\% & 60.8 & 87.2 & 97.6 & 98.5\% & 49.7 & 49.9 & 92.1 & 99.3\% \\
      DeepSeek~V4.1~Flash & 50.6 & 46.3 & 45.6 & 95.1 & 97.9\% & 86.2 & 100\% & 65.5 & 88.6 & 98.3 & 99\% & 47.4 & 46.6 & 91.2 & 97\% \\
      Doubao~Seed~2.1~Pro & 32.8 & 27.1 & 29.2 & 68.6 & 69.5\% & 84.2 & 99.5\% & 64.6 & 88.7 & 98.8 & 99.5\% & 40.6 & 38.4 & 76.5 & 84.7\% \\
      \midrule
      Text2CAD & \nodash & \nodash & \nodash & \nodash & \nodash & 12.9 & 91\% & 29.4 & 13.4 & 97.7 & 98\% & \nodash & \nodash & \nodash & \nodash \\
      \textsc{Cadrille} & \nodash & \nodash & \nodash & \nodash & \nodash & \nodash & \nodash & \nodash & \nodash & \nodash & \nodash & 21.7 & 16.6 & 74.5 & 77\% \\
      \textsc{CAD-Coder} & \nodash & \nodash & \nodash & \nodash & \nodash & \nodash & \nodash & \nodash & \nodash & \nodash & \nodash & 7.1 & 4.3 & 20.3 & 20.7\% \\
      \bottomrule
  \end{tabular}
\end{table}

\section{Experiments}
\label{sec:experiments}

\paragraph{Evaluation subsets and models.}
We evaluate two model groups on \sysname{} using a shared subset of 100 cases per task.
The first comprises ten general-purpose models: one text-only LLM and nine MLLMs.
All ten are evaluated on \Text{}, while the MLLMs are also evaluated on \Image{} and \TextImage{}.
The second comprises three domain-specific parametric CAD baselines: Text2CAD~\citep{khan2024text2cad}, \textsc{Cadrille}~\citep{kolodiazhnyi2026cadrille}, and \textsc{CAD-Coder}~\citep{doris2025cadcoderopensourcevisionlanguagemodel}.
We evaluate their released checkpoints under the I/O settings in Table~\ref{tab:models}.
Cases are randomly sampled for \Text{} and \TextImage{}; for \Image{}, we select the 100 highest-complexity cases to avoid saturation on simple inputs.
All general-purpose models use their highest supported reasoning level.
On \Image{} and \TextImage{}, general-purpose models may correct invalid outputs; we report the final outputs.
Invalid outputs remain in the evaluation denominators and receive zero bucket scores.
Table~\ref{tab:models} lists all models, parameter counts, and baseline I/O contracts.

\paragraph{Output formats per task.}
Each of the three \sysname{} tasks is evaluated on the subset of the four output formats (minimal JSON, OpenSCAD, CadQuery and Three.js) appropriate for its inputs and targets, giving seven task--format combinations in total.
\Text{} uses minimal JSON and OpenSCAD: minimal JSON matches the source construction format, while OpenSCAD is a higher-level, more readable CSG language.
\Image{} uses OpenSCAD, CadQuery and Three.js, dropping minimal JSON, which is too limited to express multi-part assemblies.
\TextImage{} uses only OpenSCAD and CadQuery, dropping Three.js as well, since its triangulated meshes are hard to decompose into per-part solids.
The per-format results below support these choices.

\begin{figure}[!t]
  \centering
  \includegraphics[width=\linewidth]{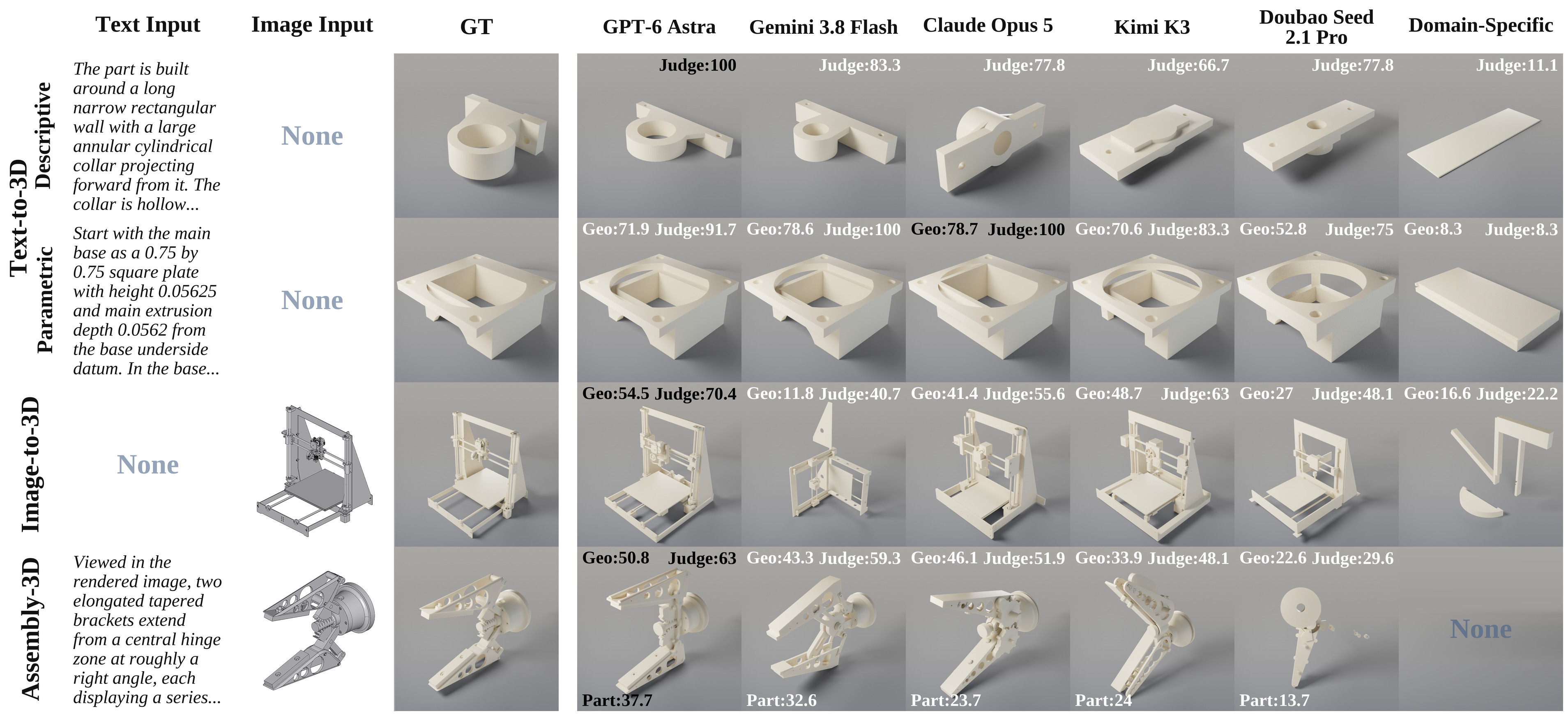}
  \caption{\textbf{Qualitative reconstructions across the three
  \sysname{} tasks.}
  Rows are tasks, with \Text{} split into descriptive and parametric.
  Five representative general-purpose models produce OpenSCAD, while
  Text2CAD on \Text{} and \textsc{Cadrille} on \Image{} are shown in
  their native formats.
  Each output is annotated with its per-case bucket scores,
  with the best in each row in black.  ``None'' marks an
  inapplicable input or baseline, and ``Part:~\nodash'' a prediction
  that failed the \emph{Part} fidelity gate.}
  \label{fig:qual-main}
\end{figure}

\begin{figure}[t]
  \centering
  \includegraphics[width=\linewidth]{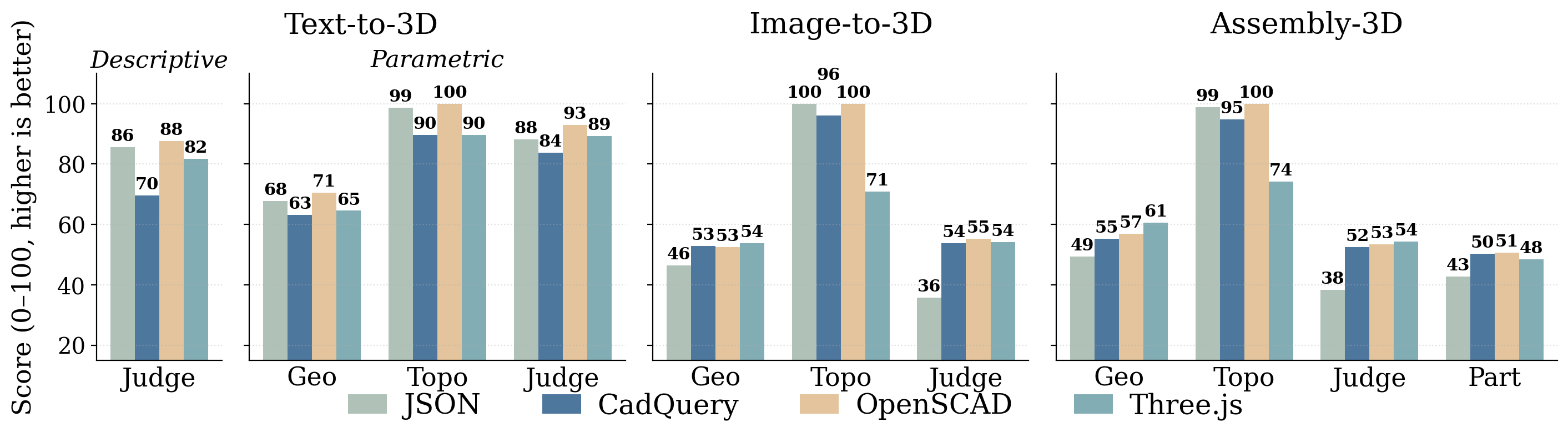}
  \caption{\textbf{Gemini~3.8~Flash bucket scores across all tasks and output
  formats.} \Text{} is split into descriptive and parametric.}
  \label{fig:format-dim-effects}
\end{figure}


\subsection{Main results}
\label{sec:exp_compare}

Table~\ref{tab:main-results} reports the cross-format average bucket scores of all evaluated models on the three \sysname{} tasks; all submetrics are reported in Appendix~\ref{app:aggregate_tables}.
Figure~\ref{fig:qual-main} shows qualitative reconstructions for five representative general-purpose models plus the applicable domain-specific baseline across the three tasks; a larger gallery with additional cases is in Figure~\ref{fig:qual-all}.

\paragraph{Across models and tasks.}
\label{sec:cmp_model_task}
GPT-6~Astra achieves the highest mean Score on all three tasks.
On \Text{}, the remaining models are tightly clustered: seven score between \(80.1\) and \(82.4\), while Doubao~Seed~2.1~Pro (\(79.2\)) and GLM-5.3-Flash (\(76.4\)) fall noticeably behind.
On \Image{} and \TextImage{}, Claude~Opus~5 and Gemini~3.8~Flash form a second tier behind GPT-6~Astra; Kimi~K3, Grok~4.6, Qwen3.8-Max and GLM-5.3-Flash occupy the middle, while DeepSeek~V4.1~Flash and especially Doubao~Seed~2.1~Pro score markedly lower.
Domain-specific models fall behind even on their native tasks and formats.
Scores are lower and more dispersed on the two assembly tasks than on the single-part task (Table~\ref{tab:main-results}, Figure~\ref{fig:teaser}).
This pattern is consistent with the additional challenges of assembly reconstruction, although the tasks differ in inputs, cases and score components.

\paragraph{Across output formats and evaluation metrics.}
\label{sec:cmp_format_dim}
To compare the four output formats, Figure~\ref{fig:format-dim-effects} reports Gemini~3.8~Flash across all four formats on every task.
For this model, OpenSCAD offers a strong balance across the applicable buckets, with no clear weakness.
JSON, in contrast, is clearly the weakest on the assembly tasks, confirming our earlier analysis.

CadQuery and Three.js are close to OpenSCAD on the overall \emph{Geo} and \emph{Judge} buckets, but fall behind for different reasons.
Models often produce invalid CadQuery programs that fail to run, especially weaker models (per-format \emph{Valid} in Appendix~\ref{app:aggregate_tables}), thereby lowering their scores across all metrics.
Three.js instead has no solid kernel and builds triangulated surfaces, so it scores poorly on \emph{Topo} and \emph{Part}: such meshes are not guaranteed watertight and do not yield clean per-part solids for matching.

\subsection{Result details}
\label{sec:exp_details}

The headline \emph{Judge} and \emph{Part} buckets in Table~\ref{tab:main-results} each average several submetrics into a single score.
In this section, we unpack both buckets and analyze their submetrics.

\begin{table}[t]
  \centering
  \caption{\textbf{The \emph{Judge} bucket in detail.}
  \textbf{(a)} Gemini~3.8~Flash's submetrics averaged over the main-experiment
  formats, with \Text{} split into descriptive and parametric;
  ``\nodash'' marks an inapplicable submetric, and qualitative cases are
  shown in Figure~\ref{fig:judge-details-qual}.
  \textbf{(b)} Agreement with human annotators: for \Text{}, the rate at
  which an annotator and the MLLM Judge select the same multiple-choice
  option; for \Image{} and \TextImage{}, Spearman's \(\rho\) against the
  mean human rating (\(95\%\) CI).}
  \label{tab:judge}
  \footnotesize
  \setlength{\tabcolsep}{3pt}
  \renewcommand{\arraystretch}{1.05}
  \begin{subtable}[t]{0.54\textwidth}
    \centering
    \caption{Judge submetrics}
    \label{tab:judge-details}
    \begin{tabular}{@{}lccccc@{}}
      \toprule
      Task & QA-S & QA-P & J-Geo & J-Aes & J-Sem \\
      \midrule
      \Text{} (Desc.) & 0.98 & \nodash & \nodash & \nodash & 0.75 \\
      \Text{} (Param.) & 0.89 & 0.91 & \nodash & \nodash & \nodash \\
      \Image{} & \nodash & \nodash & 0.43 & 0.54 & 0.65 \\
      \TextImage{} & \nodash & \nodash & 0.43 & 0.52 & 0.64 \\
      \bottomrule
    \end{tabular}
  \end{subtable}
  \hfill
  \begin{subtable}[t]{0.44\textwidth}
    \centering
    \caption{Judge vs.\ human annotators}
    \label{tab:judge-human}
    \begin{tabular}{@{}lc@{}}
      \toprule
      Task & QA-S agreement \\
      \midrule
      \Text{} & 68.3\% \\
      \midrule
      Task & Spearman \(\rho\) [95\% CI] \\
      \midrule
      \Image{}     & 0.924 [.70,\,.98] \\
      \TextImage{} & 0.921 [.69,\,.98] \\
      \bottomrule
    \end{tabular}
  \end{subtable}
\end{table}

\begin{figure}[!t]
  \centering
  \includegraphics[width=\linewidth]{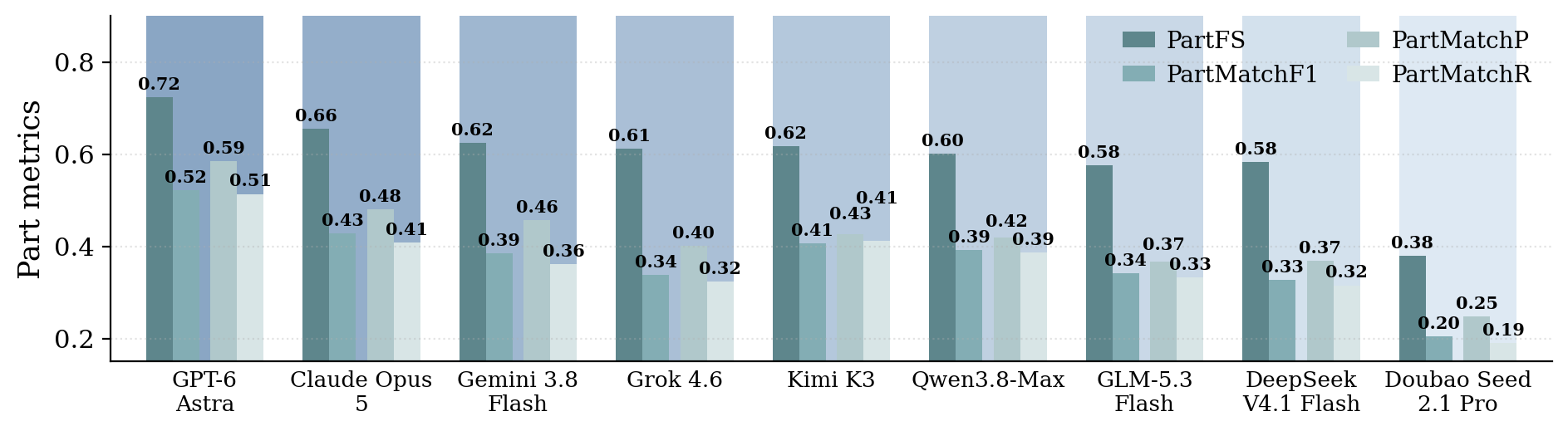}
  \caption{\textbf{Part metrics on \TextImage{}.}
  The \emph{Part} bucket submetrics PartFS and PartMatchF1, with
  PartMatchP and PartMatchR, averaged across CadQuery and OpenSCAD.
  Qualitative cases are shown in Figure~\ref{fig:textimage-part-detail-b}.}
  \label{fig:textimage-part-detail-a}
\end{figure}

\paragraph{Judge bucket details.}
Table~\ref{tab:judge-details} reports Gemini~3.8~Flash's Judge submetrics using the same task-specific format averages as Table~\ref{tab:main-results}.
\textbf{(i)} QA-S is lower under parametric than descriptive specifications (\(0.89\) vs.\ \(0.98\)), while parametric QA-P reaches \(0.91\). These results highlight that the remaining challenges in parametric generation extend beyond numerical accuracy to satisfying semantic constraints.
\textbf{(ii)} On \Image{} and \TextImage{}, \emph{J-Sem} reaches \(0.65\) and \(0.64\), respectively, while \emph{J-Geo} rounds to \(0.43\) on both tasks. For Gemini~3.8~Flash, these ratings suggest stronger agreement on object identity than on precise shape and proportions.
Table~\ref{tab:judge-human} shows overall consistency between the MLLM Judge and human judgments: QA-S agreement is \(68.3\%\) on predictions (human accuracy on ground truth: \(80.0\%\)), while visual scores correlate strongly with mean human ratings on \Image{} and \TextImage{} (Spearman \(\rho\ge0.92\)); see Appendix~\ref{app:judge_human_study} for study details.

\paragraph{Part metric details.}
Figure~\ref{fig:textimage-part-detail-a} reports the Part metrics on \TextImage{}---PartFS and PartMatchF1, with PartMatchP and PartMatchR---averaged across CadQuery and OpenSCAD.
\textbf{(i)} PartFS, per-part geometric fidelity of matched parts, reaches only \(\approx 0.72\) for the strongest model, GPT-6~Astra, and drops sharply for weaker ones. This is consistent with the Judge submetrics on the same task (Table~\ref{tab:judge-details}), where semantic alignment (\emph{J-Sem}) exceeds geometric alignment (\emph{J-Geo}): models capture overall shape semantics more reliably than reconstruct fine-grained per-part geometry.
\textbf{(ii)} PartMatchF1 reaches only \(\approx 0.52\) even for the best model, with PartMatchP and PartMatchR both near \(0.59\) and \(0.51\): only about half of the predicted and ground-truth parts are successfully matched. Their ratio further reveals a mismatch between the predicted and ground-truth part counts.
Current models thus recover neither the part count nor the per-part geometry of the assembly.

\FloatBarrier

\section{Conclusion}
\label{sec:conclusion}

We introduced \sysname{}, a unified benchmark for executable parametric 3D generation across text, image and assembly tasks.
\sysname{} clearly distinguishes models with different levels of parametric 3D generation capability.
Across tasks and output formats, the main bottleneck is not producing runnable and visually plausible outputs, but translating specifications into precise geometry and part structure.
A current limitation is that, while the appendix includes a preliminary multi-turn experiment (Table~\ref{tab:image-agent-loop}), \sysname{} does not yet systematically evaluate general-purpose coding agents such as Codex and Claude Code; extending it to such agents, which execute and revise their programs over multiple turns, is left to future work.

\FloatBarrier

\bibliography{references}
\bibliographystyle{iclr2027_conference}

\appendix
\renewcommand{\thefigure}{A\arabic{figure}}
\renewcommand{\thetable}{A\arabic{table}}
\setcounter{figure}{0}
\setcounter{table}{0}
\section*{Overview}
\label{app:appendix}

This appendix collects implementation details and supplementary analyses that support the main benchmark description.
\emph{Dataset processing details} documents the preprocessing, filtering and annotation steps and the statistics of the resulting sets; \emph{Experimental setup} lists the evaluated models and their inference setup; \emph{Evaluation details} covers mesh preprocessing, score aggregation, QA bank construction and \TextImage{} \emph{Part} metrics.
\emph{Additional analyses and experiments} reports human validation of the MLLM Judge, qualitative case studies of the Judge and Part submetrics, three diagnostic analyses, and a multi-turn agentic workflow.
\emph{Full submetric results} reports every submetric for the three tasks under each output format, and \emph{Qualitative visualizations} presents qualitative outputs from the evaluated models.

\section{Dataset processing details}
\label{app:dataset_details}

This section details dataset preprocessing, filtering and annotation, followed by statistics and annotation examples.

\newcommand{\dsFigScale}{0.90}

\subsection{Pipeline implementation}
\label{app:pipeline_details}
\label{app:source_preprocess}
\label{app:filtering_details}
\label{app:annotation_details}

\paragraph{Text2CAD preprocessing.}
Preprocessing reduces the \(176{,}017\) Text2CAD records~\citep{khan2024text2cad} to the candidate pool for filtering.
We discard records that cannot be evaluated at all --- missing programs, zero-depth extrusions and empty shapes --- and rank the survivors by a geometric complexity score over the modeling-operation sequence, namely the number of sketch and extrude operations and the face count, keeping the top-ranked cases as candidates.
The Fusion~360 Gallery assemblies~\citep{willis2022joinable} already carry multi-part structure and skip this preselection.

\paragraph{Near-duplicate removal.}
\label{app:dedup_details}
The candidates retained after the review MLLM still contain many visually similar parts and assemblies, which we remove by DINOv2~\citep{oquab2024dinov2} feature matching over render embeddings, applied identically to both sources.
Each candidate is embedded by the CLS token of DINOv2's last hidden state on its render, L2-normalized so that cosine similarity is a dot product.
We then walk the pairs with similarity \(\ge \tau\) in descending order and, for each pair still alive, drop the endpoint with the higher average similarity to the remaining alive set, since the case that looks more like everyone else is the more redundant one.

\begin{figure}[!p]
  \centering
  \includegraphics[width=\dsFigScale\linewidth]{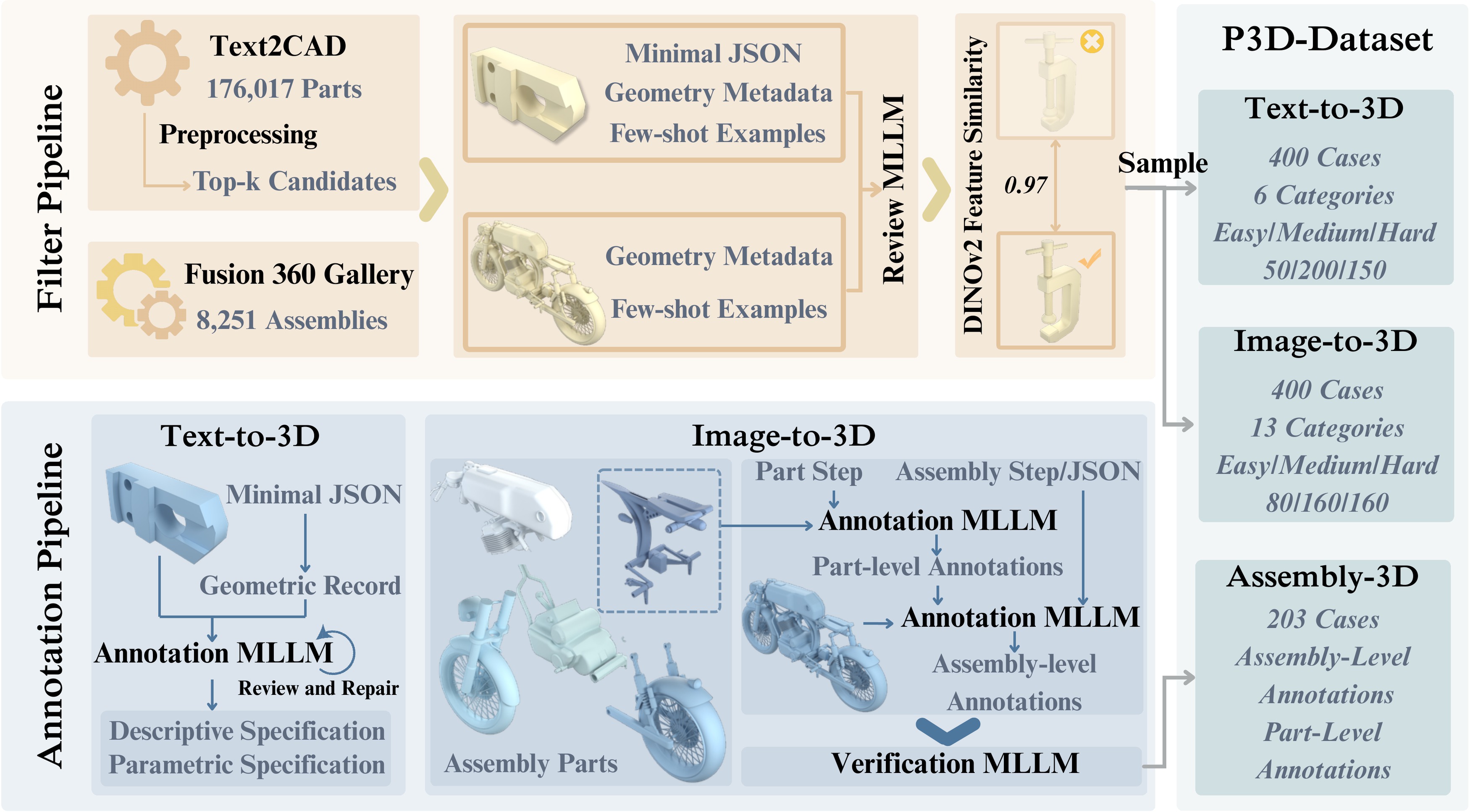}
  \caption{\textbf{Overview of our P3D-Dataset.} The \emph{filter pipeline}
  draws from the Text2CAD and Fusion~360 Gallery data sources, filters
  candidates with a review MLLM, removes near-duplicates, and samples
  \(400\) \Text{} and \(400\) \Image{} cases with category and complexity
  distributions. The \emph{annotation pipeline} then annotates the filtered
  \Text{} data with an annotation MLLM into descriptive (desc) and parametric
  (param) text specifications; for the \Image{} data, both parts and
  assemblies are labeled by an annotation MLLM and checked by a verification
  MLLM, yielding the \(203\)-case \TextImage{} set.}
  \label{fig:data-pipeline}
\end{figure}

\paragraph{Text annotation.}
As shown in the annotation pipeline of Figure~\ref{fig:data-pipeline}, the two \Text{} specifications are not written from the raw source program.
We first parse the source minimal JSON into a structured geometric record that makes the part's sketches, extrusions and bounding boxes explicit, and pass it with the render to the annotation MLLM (GPT-5.5), so that both specifications are grounded in named geometric quantities rather than inferred from the render.
The record also backs the static validator that closes the review-and-repair loop --- a case is admitted only after every number and feature in its text checks out against it --- and keeps the two specifications distinct, allowing no numerical parameters in the descriptive one and only recorded features in the parametric one. Both specifications are written as natural-language design descriptions.

\subsection{Dataset statistics and annotation examples}
\label{app:dataset-statistics}
\label{app:annotation-examples}

\begin{figure}[!p]
  \centering
  \begin{subfigure}[t]{\dsFigScale\linewidth}
    \centering
    \includegraphics[width=\linewidth]{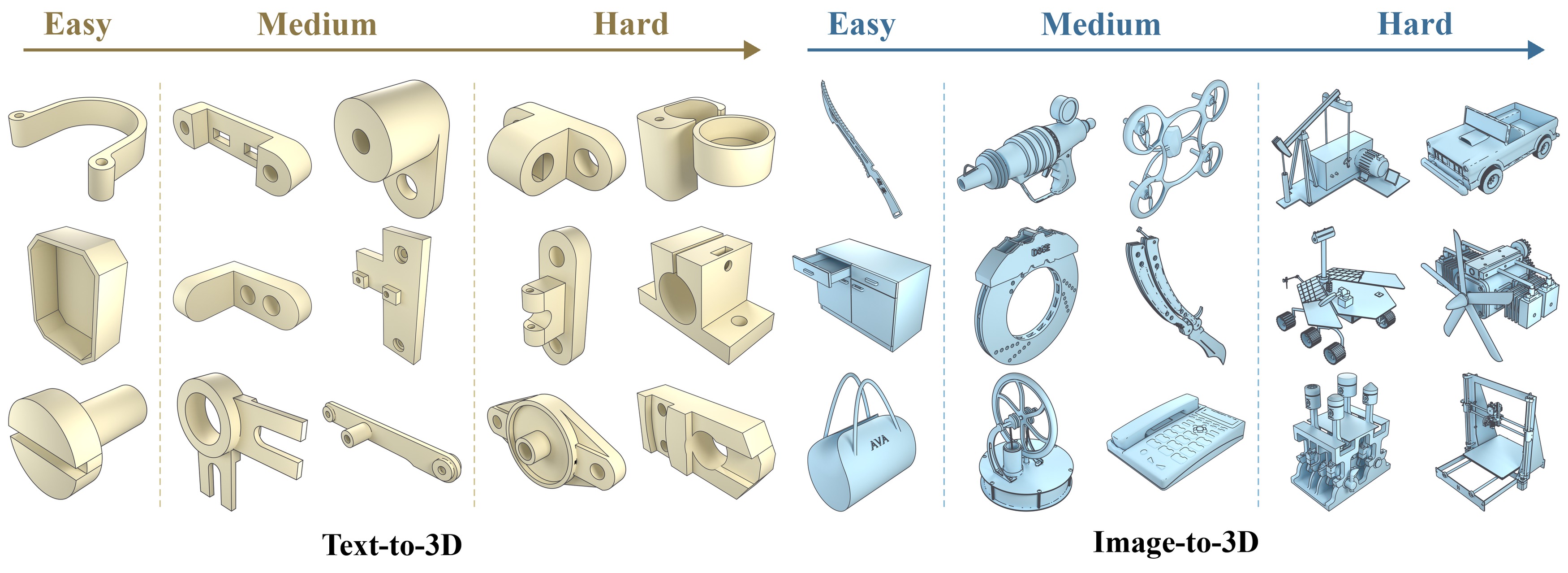}
    \caption{Example cases at the easy, medium and hard complexity levels.}
    \label{fig:dataset-stats-gallery}
  \end{subfigure}
  \vspace{2pt}
  \begin{subfigure}[t]{\dsFigScale\linewidth}
    \centering
    \includegraphics[width=\linewidth]{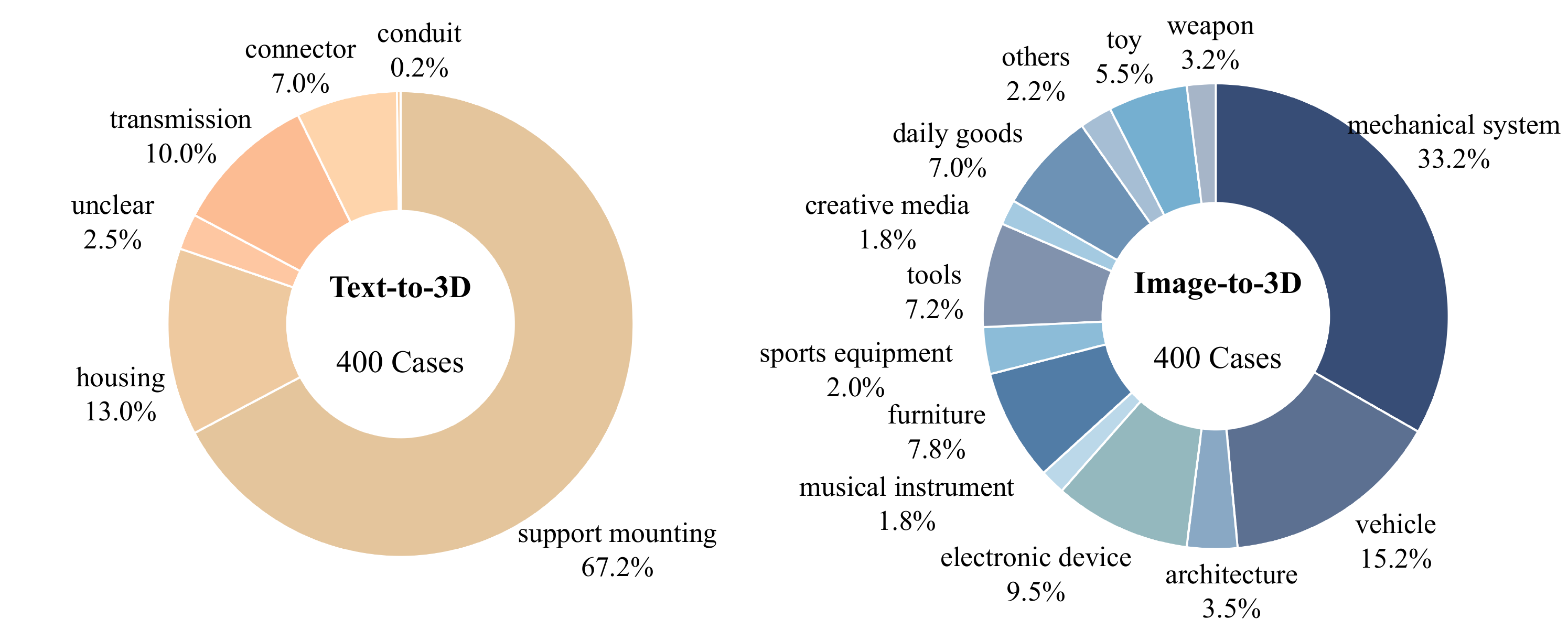}
    \caption{Semantic category distribution of the \Text{} and \Image{} sets.}
    \label{fig:dataset-stats-donut}
  \end{subfigure}
  \caption{\textbf{The \Text{} and \Image{} 400-case filtered sets.}}
  \label{fig:dataset-stats}
\end{figure}

\begin{figure}[t]
  \centering
  \includegraphics[width=\linewidth]{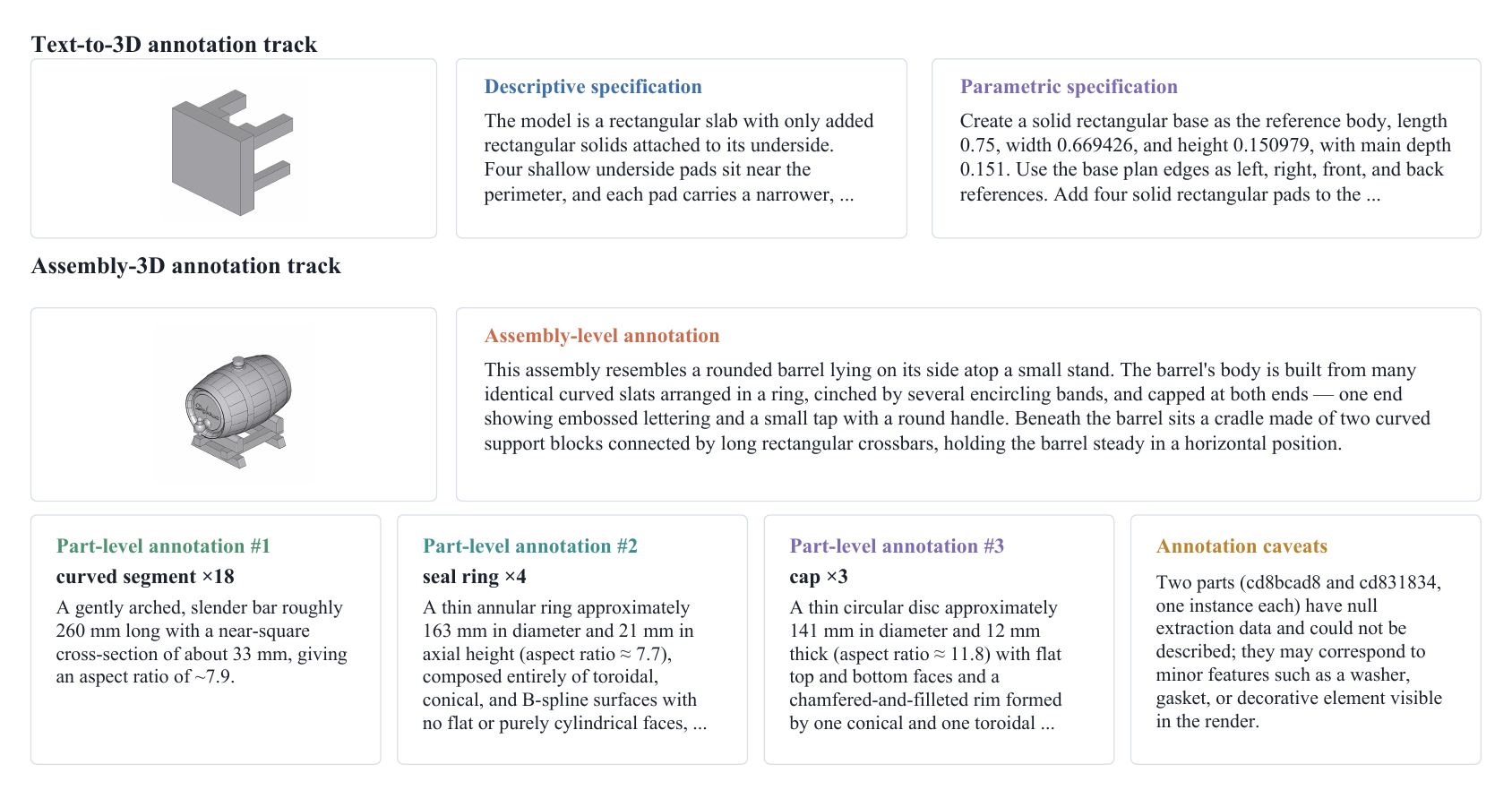}
  \caption{\textbf{Example annotations from the two data sources.}  Top: a
  retained Text2CAD program with its source render, converted into the
  descriptive and parametric specifications used by \Text{}.  Bottom: a
  Fusion~360 assembly render with the verified assembly-level caption and
  per-part descriptions used by \TextImage{}.  Text boxes show verbatim
  excerpts of the released input fields; the benchmark prompts use the
  full fields.}
  \label{fig:annotation-examples}
\end{figure}

Figure~\ref{fig:dataset-stats} summarizes the two \(400\)-case filtered sets.
The semantic category distribution (Figure~\ref{fig:dataset-stats-donut}) is long-tailed for both: six part-level categories for \Text{}, dominated by support mounting, and thirteen assembly-level categories for \Image{}, led by mechanical systems and vehicles.
Figure~\ref{fig:dataset-stats-gallery} shows example cases at the easy, medium and hard complexity levels.
Complexity-balanced sampling keeps all three tiers populated: the \Text{} set splits \(50/200/150\) over easy, medium and hard, and the \Image{} set \(80/160/160\).
The \TextImage{} annotation pipeline then drops \Image{} assemblies that exceed the deduplicated-part cap or fail MLLM verification, leaving \(203\). These are the complete dataset sizes; the main experiments use a fixed \(100\)-case subset of each task.
Figure~\ref{fig:annotation-examples} shows the resulting annotations for both tracks.

\section{Experimental setup}
\label{app:experimental_setup}

Table~\ref{tab:models} lists the evaluated models by type, with parameter
counts and each domain-specific baseline's input--output setting used here.
All general-purpose LLMs and MLLMs use their highest supported reasoning level.
Domain-specific models run from their released checkpoints under those
settings, and generated programs are executed in their native
toolchains---OpenSCAD~\citep{openscad}, CadQuery~\citep{cadquery} and
Three.js~\citep{threejs}.
We use Cadrille's image-conditioned mode.
Local evaluation used Ubuntu~22.04 with dual Intel Xeon Silver~4314 CPUs,
512~GB RAM and eight NVIDIA RTX~A6000 GPUs, under Python~3.12.2,
OpenSCAD~2021.01 and CadQuery~2.7.0.

LLM-based assistants supported code implementation and debugging, experiment execution and analysis, literature retrieval, and manuscript drafting and editing; the authors reviewed the outputs and take responsibility for the final work.

\begin{table}[!t]
  \centering
  \caption{\textbf{Models evaluated in \sysname{}, grouped by model type.}  Size is
  the developer-reported parameter count, with active parameters per token for MoE;
  ``\nodash'' marks closed-weight models with undisclosed counts, and
  domain-specific sizes refer to the backbone. DeepSeek's count includes its backbone and conditional memory. Markers give each
  domain-specific baseline's input $\to$ output setting used here:
  \textsuperscript{$\ast$}\,image $\to$ CadQuery,
  \textsuperscript{$\S$}\,text $\to$ minimal JSON.
  GLM-5.3-Flash is the multimodal variant evaluated on all three tasks;
  GLM-5.3 is the text-only variant evaluated only on \Text{}.}
  \label{tab:models}
  \setlength{\tabcolsep}{6pt}
  \renewcommand{\arraystretch}{1.15}
  \small
  \resizebox{\linewidth}{!}{%
  \begin{tabular}{@{}llc@{}}
    \toprule
    Model type & Model & Size (total\,/\,active) \\
    \midrule
    \multirow{9}{*}{Multimodal LLM}
      & Claude~Opus~5~\citep{anthropic2026opus5}        & \nodash                 \\
      & Gemini~3.8~Flash~\citep{google2026gemini38flash} & \nodash                 \\
      & GPT-6~Astra~\citep{openai2026gpt6astra} & \nodash                 \\
      & Qwen3.8-Max~\citep{qwen2026qwen38max}          & 2.4\,T\,/\,\nodash      \\
      & GLM-5.3-Flash~\citep{zai2026glm53flash}        & 320\,B\,/\,18\,B        \\
      & Doubao~Seed~2.1~Pro~\citep{bytedance2026seed21} & \nodash                 \\
      & Kimi~K3~\citep{moonshot2026kimik3}             & 2.8\,T\,/\,104\,B       \\
      & Grok~4.6~\citep{xai2026grok46}                 & \nodash                 \\
      & DeepSeek~V4.1~Flash~\citep{deepseekai2026deepseekv41flash} & 748\,B\,/\,8--16\,B \\
    \midrule
    Text-only LLM
      & GLM-5.3~\citep{zai2026glm53}                   & 753\,B\,/\,\nodash      \\
    \midrule
    \multirow{3}{*}{Domain-specific model}
      & \textsc{Cadrille}\textsuperscript{$\ast$}~\citep{kolodiazhnyi2026cadrille}  & 2\,B (Qwen2-VL-2B) \\
      & \textsc{CAD-Coder}\textsuperscript{$\ast$}~\citep{doris2025cadcoderopensourcevisionlanguagemodel} & 13\,B (Vicuna-13B) \\
      & Text2CAD\textsuperscript{$\S$}~\citep{khan2024text2cad}  & 363\,M (BERT-Large + decoder) \\
    \bottomrule
  \end{tabular}}
\end{table}

\section{Evaluation details}
\label{app:evaluation_details}

\paragraph{Bucket score aggregation.}
\label{app:bucket_details}
\label{app:invalid_missing}

Invalid predictions remain in the evaluation denominators, with zero normalized sub-metrics and bucket scores and worst-case raw errors (\(0.01\) for CD).
For otherwise valid predictions, unavailable IoU is omitted from per-case Geo and the standalone IoU mean; measured zeros and invalid-output penalties are retained.
Sub-metrics are mapped to \([0,1]\), with \(1\) best: visual-Judge ratings use \((s-1)/9\) for \(s\in[1,10]\), and CD uses \(\max(0,\,1-\mathrm{CD}/0.01)\); reported CD is capped at \(0.01\) before averaging.
Each bucket averages its applicable normalized sub-metrics equally, except parametric \Text{} Judge, which uses \((\text{QA-S}+2\,\text{QA-P})/3\) to reflect its four QA-S and eight QA-P questions.
Geo is computed per prediction before averaging over cases within each output format; reported bucket scores weight the task-supported formats equally.
Part measurements rejected by the fidelity gate are excluded from Part means.

The headline Score averages descriptive Judge, parametric Geo and parametric Judge on \Text{}; Geo and Judge on \Image{}; and Geo, Judge and Part on \TextImage{}, then scales each mean to $0$--$100$.

\paragraph{Mesh normalization, alignment and IoU.}
\label{app:mesh_alignment_details}

We center both meshes and, when enabled, normalize their scale to the unit cube; parametric \Text{} retains absolute scale because dimensions are specified in the input.
Alignment minimizes bidirectional CD over axis permutations and sign flips.
Candidates within \(5\%\) of the minimum are treated as tied, and the rotation closest to identity is retained to preserve the generated orientation when possible.
A search in the prediction's PCA frame is accepted only if it improves CD by at least \(10\%\).
We then refine translation and, when permitted, uniform scale with a locally bounded search.
IoU uses CSG solids (\IoUC{}$\uparrow$) for single-part \TextParam{} outputs and voxel grids (\IoUV{}$\uparrow$) for \Image{} and \TextImage{} assemblies.
Voxel IoU is more robust to non-manifold assembly geometry because it requires a closed surface rather than a manifold solid.
Both variants require the predicted and ground-truth meshes to be free of open edges.
For an otherwise valid prediction with an open surface, IoU is undefined
rather than a measured zero, so it is omitted from that case's Geo mean
and the standalone IoU mean; NoOE reports the open-edge condition separately.
Invalid-output penalties and measured zeros are retained as described above.

\paragraph{QA bank construction.}
\label{app:judge_metric_details}

Both banks are synthesized offline by Gemini~3.1~Pro, once per case and then frozen; questions are limited to the given text specification, with the ground-truth program and render used only to verify answers.
Each question has five options, including ``none of the above'', and some questions deliberately use that answer so that the grader cannot succeed by always choosing the nearest concrete value.
QA-S excludes exact numeric values, whereas QA-P mainly asks for derived quantities that require combining measurements or applying a geometric formula rather than copying a stated parameter; its distractors reflect plausible incorrect derivations.
Finally, the same model re-derives every answer from the ground truth and rewrites any question it finds incorrect.
At evaluation time, Gemini~3.8~Flash answers the frozen questions to score each prediction.

\paragraph{\TextImage{} \emph{Part} metrics.}
\label{app:textimage_part_metrics}

Parts share one alignment transform per side and are never rescaled, preserving differences in relative part size.
We sample \(2048\) surface points per part with a fixed seed.
Each candidate pair is centroid-aligned and takes the best of the \(24\) proper cube rotations under bidirectional CD.
Hungarian matching uses a coverage F-score at \(\tau_g=0.03\,\mathrm{diag}_g\), and a pair is successful at \(F_{\min}=0.7\).
A rotation- and translation-invariant fingerprint first deduplicates repeated instances of the same body.

\begin{table}[!p]
  \centering
  \caption{\textbf{Human validation of the MLLM Judge.}
  \textbf{(a)} QA-S agreement with the reference answer key and with the
  Judge's answer on the prediction.
  \textbf{(b)} Pearson \(r\) and Spearman \(\rho\) between the visual Judge
  and mean human ratings over \(10\) cases per task; \emph{Composite} is
  the mean of the three axes and \emph{Pooled} uses all \(20\) cases.}
  \label{tab:appendix-judge-human}
  \footnotesize
  \setlength{\tabcolsep}{5pt}
  \renewcommand{\arraystretch}{1.05}
  \begin{subtable}[t]{\linewidth}
    \centering
    \caption{QA metric, on \Text{} QA-S}
    \begin{tabular}{@{}lcc@{}}
      \toprule
      Metric & Reference & Prediction \\
      \midrule
      QA-S agreement & 80\% & 68.3\% \\
      \bottomrule
    \end{tabular}
  \end{subtable}

  \vspace{5pt}

  \begin{subtable}[t]{\linewidth}
    \centering
    \caption{Visual judge, on \Image{} and \TextImage{}}
    \begin{tabular}{@{}lcccccc@{}}
      \toprule
      & \multicolumn{2}{c}{\Image{}} & \multicolumn{2}{c}{\TextImage{}}
      & \multicolumn{2}{c}{Pooled} \\
      \cmidrule(lr){2-3} \cmidrule(lr){4-5} \cmidrule(l){6-7}
      Axis & \(r\) & \(\rho\) & \(r\) & \(\rho\) & \(r\) & \(\rho\) \\
      \midrule
      \emph{J-Geo} & 0.824 & 0.870 & 0.850 & 0.812 & 0.838 & 0.860 \\
      \emph{J-Aes} & 0.930 & 0.858 & 0.870 & 0.821 & 0.888 & 0.853 \\
      \emph{J-Sem} & 0.928 & 0.815 & 0.934 & 0.911 & 0.930 & 0.894 \\
      Composite    & 0.961 & 0.924 & 0.951 & 0.921 & 0.952 & 0.931 \\
      \bottomrule
    \end{tabular}
  \end{subtable}
\end{table}

\begin{figure}[!p]
  \centering
  \includegraphics[width=\linewidth]{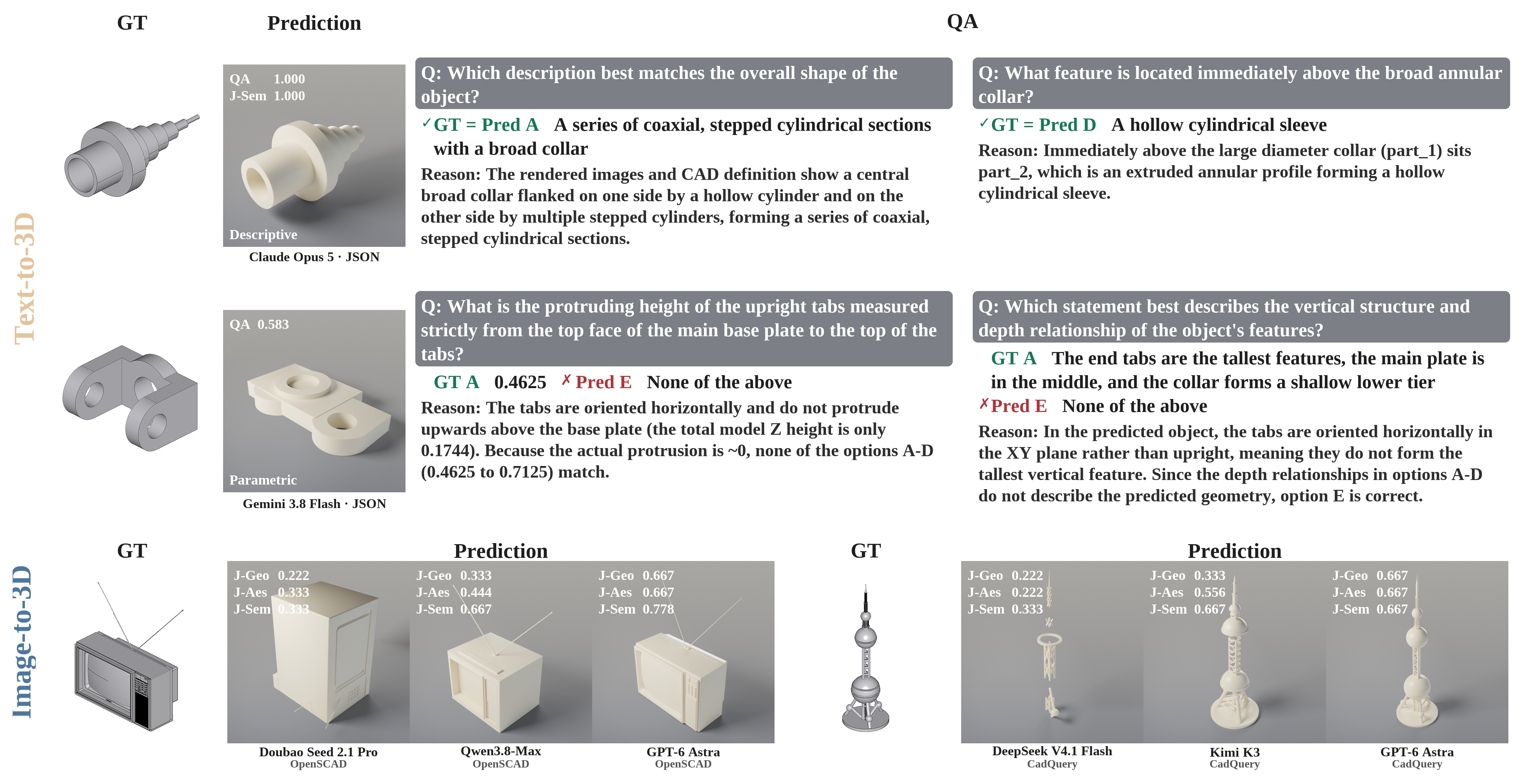}
  \caption{\textbf{Qualitative cases showing Judge submetrics.}
  \emph{(top)} One descriptive and one parametric specification example,
  each with its QA results and reasons.
  \emph{(bottom)} Two \Image{} cases with their visual Judge submetrics
  \emph{J-Geo}/\emph{J-Aes}/\emph{J-Sem}.}
  \label{fig:judge-details-qual}
\end{figure}

\begin{figure}[!p]
  \centering
  \includegraphics[width=\linewidth]{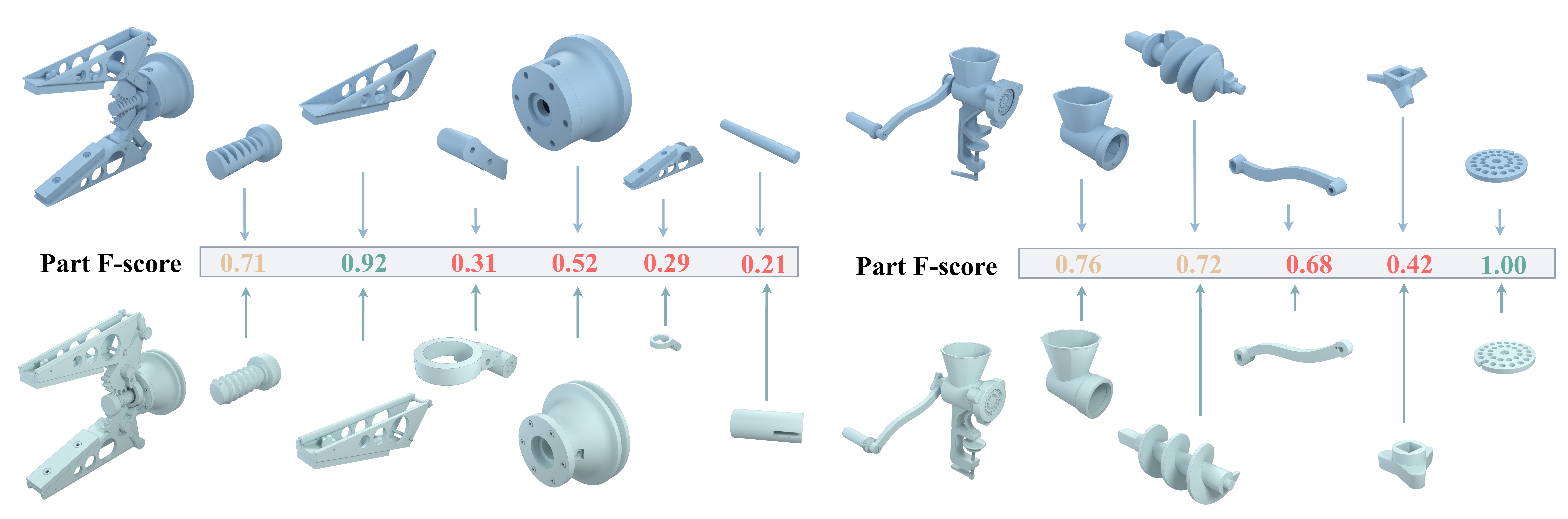}
  \caption{\textbf{Qualitative cases showing part-match results.}
  Selected ground-truth parts (top) from two complex assemblies are
  paired with predicted parts (bottom) by Hungarian assignment, with each pair annotated
  by its per-part F-score \(F\) at the $3\%$ tolerance: green for \(F\geq 0.9\), yellow for
  \(0.7\leq F<0.9\), and red for rejected pairs with \(F<0.7\).}
  \label{fig:textimage-part-detail-b}
\end{figure}

\section{Additional analyses and experiments}
\label{app:additional_analyses}
\label{sec:exp_extra}

We present human validation of the Judge, qualitative Judge and Part examples, and analyses of cost, decomposition fidelity, output validity and multi-turn feedback.

\subsection{Human validation of the MLLM Judge}
\label{app:judge_human_study}

We validate the \emph{Judge} components that humans can reproduce: QA-S for \Text{}, and \emph{J-Geo}, \emph{J-Aes} and \emph{J-Sem} for \Image{} and \TextImage{}; QA-P is omitted because it requires deriving quantities from predicted source code.
For each task, human raters assess the same \(10\) cases. Visual-Judge cases are score-stratified; QA-S items span correct and incorrect Judge answers.

\paragraph{\Text{}: QA-S.}
Given the input specification and the ground-truth or predicted render, annotators answer the same QA-S items as the Judge.
Their responses agree with the reference answer key in \(80.0\%\) of trials and with the Judge's answer on the prediction in \(68.3\%\) (Table~\ref{tab:appendix-judge-human}a).

\paragraph{\Image{} and \TextImage{}: visual judge.}
Annotators receive the same evidence and rubric as the MLLM Judge and rate \emph{J-Geo}, \emph{J-Aes} and \emph{J-Sem} on the same \(1\)--\(10\) scale.
We average the three axes per case and compare the Judge score with the mean human score using Pearson \(r\) and Spearman \(\rho\).
The composite agreement is strong on the sampled cases (Table~\ref{tab:appendix-judge-human}b): \(r=0.961,\rho=0.924\) on \Image{} and \(r=0.951,\rho=0.921\) on \TextImage{}, indicating that the MLLM Judge closely tracks human score variation and rankings on these cases.

\subsection{Qualitative Judge and Part results}
\label{app:judge_part_qualitative}

Figures~\ref{fig:judge-details-qual} and~\ref{fig:textimage-part-detail-b} provide qualitative examples of the Judge and Part metrics, respectively.  The former compares Judge submetrics across \Text{} and \Image{} cases, while the latter visualizes part correspondences and their per-part F-scores in \TextImage{}.

Across these examples, the metric values agree with the qualitative evidence: outputs that better match the reference receive higher scores, whereas visible structural, geometric, or part-matching errors receive lower scores.  This consistency supports the validity of both metrics.

\subsection{Cost-quality analysis}
\label{app:api_efficiency}

\begin{figure}[!t]
  \centering
  \includegraphics[width=\linewidth]{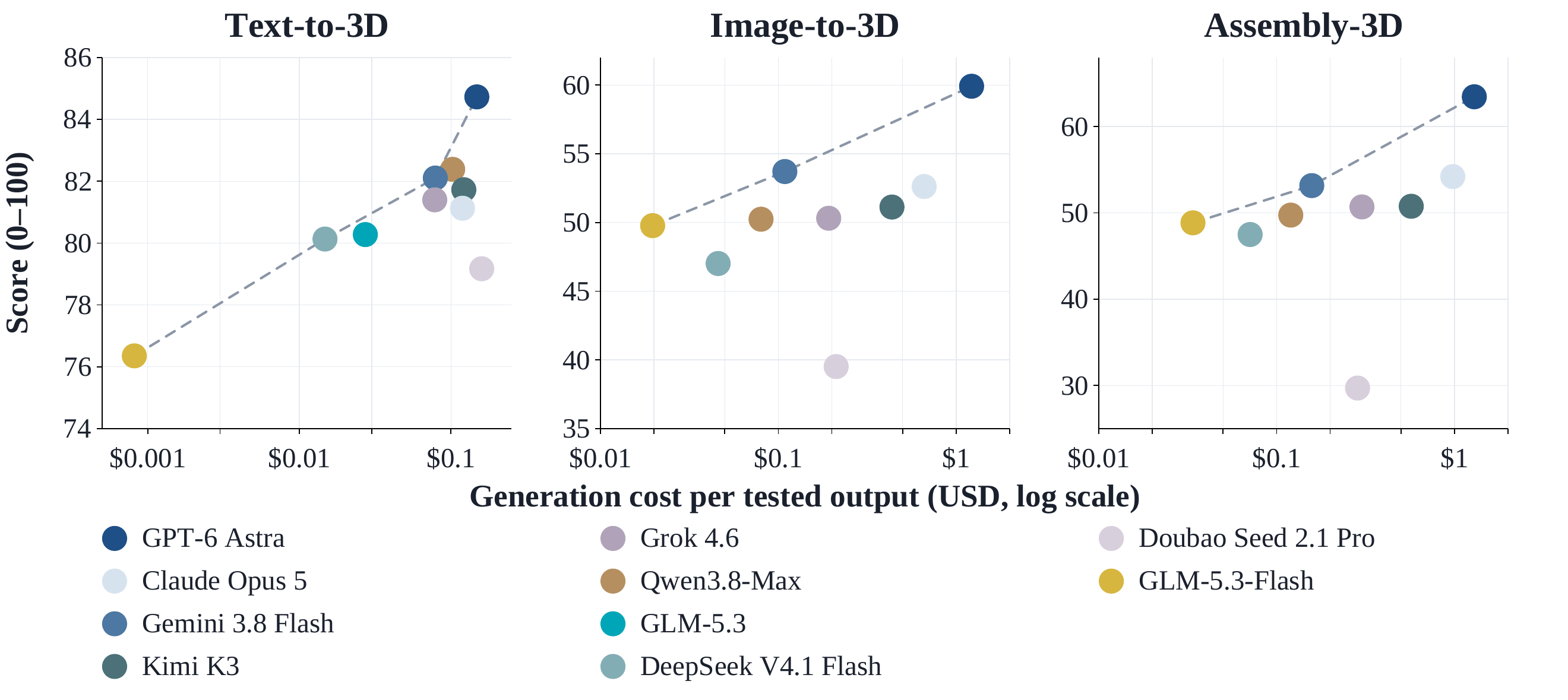}
  \caption{\textbf{Per-task \sysname{} Score versus mean API-equivalent
  generation cost per tested output (USD, log scale).} Each point represents
  a model; marker colour identifies the model, and the dashed line connects
  selected models to show the cost--quality trend.}
  \label{fig:api-efficiency-cost}
\end{figure}

Figure~\ref{fig:api-efficiency-cost} plots task-specific \sysname{}
scores against mean API-equivalent generation cost per tested output.
For each model and task, we divide total generation cost by the number of
tested outputs, including invalid ones. For \Text{}, the cost includes only
the selected generation for each output. For \Image{} and \TextImage{}, it
also includes recorded execution-feedback corrections; the correction
protocol is detailed in Appendix~\ref{sec:invalid}.

GPT-6~Astra occupies the high-quality, high-cost end across all three tasks.
Gemini~3.8~Flash offers a cost-effective option at intermediate budgets
across all three tasks. GLM-5.3-Flash offers a cost-effective alternative
across all three tasks. DeepSeek~V4.1~Flash is also cost-effective on \Text{}.

\subsection{\TextImage{} decomposition fidelity}
\label{app:textimage_supplements}

Table~\ref{tab:appendix-textimage-decomp} reports per-model decomposition
fidelity: how closely the union of the re-assembled decomposition matches
the prediction it came from, in Chamfer Distance
(\(\mathrm{D}_{\mathrm{dec}}\), lower is better) and volumetric IoU
(\(\mathrm{IoUV}_{\mathrm{dec}}\), higher is better), together with the
resulting gate exclusions.

\begin{table}[!t]
  \centering
  \caption{\textbf{Decomposition fidelity diagnostic per model and output format.}
  \emph{Valid} is the executable-prediction rate, reported as a percentage;
  \(\mathrm{D}_{\mathrm{dec}}\) and \(\mathrm{IoUV}_{\mathrm{dec}}\) are
  the predicted vs.\ re-assembled union Chamfer Distance and volumetric
  IoU, averaged over the same valid, successfully decomposed cases so
  that neither is confounded by \emph{Valid}; \emph{Excl.} counts the
  cases dropped by the fidelity gate.}
  \label{tab:appendix-textimage-decomp}
  \setlength{\tabcolsep}{4pt}
  \renewcommand{\arraystretch}{1.35}
  \small
  \resizebox{\linewidth}{!}{%
  \begin{tabular}{lcccccccc}
    \toprule
    & \multicolumn{4}{c}{CadQuery} & \multicolumn{4}{c}{OpenSCAD} \\
    \cmidrule(lr){2-5} \cmidrule(lr){6-9}
    Model & Valid\(\uparrow\) & \(\mathrm{D}_{\mathrm{dec}}\)\(\downarrow\) & \(\mathrm{IoUV}_{\mathrm{dec}}\)\(\uparrow\) & Excl.\(\downarrow\)
          & Valid\(\uparrow\) & \(\mathrm{D}_{\mathrm{dec}}\)\(\downarrow\) & \(\mathrm{IoUV}_{\mathrm{dec}}\)\(\uparrow\) & Excl.\(\downarrow\) \\
    \midrule
      GPT-6~Astra & 99\% & 0.000075 & 0.987 & 0 & 100\% & 0.000076 & 0.986 & 0 \\
      Claude~Opus~5 & 100\% & 0.000210 & 0.973 & 1 & 100\% & 0.000071 & 0.983 & 0 \\
      Gemini~3.8~Flash & 98\% & 0.000073 & 0.982 & 0 & 100\% & 0.000070 & 0.985 & 0 \\
      Kimi~K3 & 98\% & 0.000072 & 0.984 & 0 & 100\% & 0.000274 & 0.964 & 3 \\
      Grok~4.6 & 100\% & 0.000581 & 0.964 & 1 & 100\% & 0.000068 & 0.985 & 0 \\
      Qwen3.8-Max & 98\% & 0.000064 & 0.987 & 0 & 100\% & 0.000069 & 0.984 & 0 \\
      GLM-5.3-Flash & 98\% & 0.000076 & 0.983 & 0 & 100\% & 0.000069 & 0.987 & 0 \\
      DeepSeek~V4.1~Flash & 97\% & 0.000260 & 0.978 & 1 & 99\% & 0.000071 & 0.989 & 0 \\
      Doubao~Seed~2.1~Pro & 41\% & 0.000524 & 0.935 & 2 & 98\% & 0.000480 & 0.927 & 5 \\
      \bottomrule
  \end{tabular}}
\end{table}

\begin{table}[!t]
  \centering
  \caption{\textbf{First and final invalid rates on the grounded tasks.}
  Rates are reported as percentages and averaged over the output formats evaluated for each task in the
  main experiments. Final denotes the released outcome after the recorded
  corrections.}
  \label{tab:invalid-first-final}
  \setlength{\tabcolsep}{2.5pt}
  \renewcommand{\arraystretch}{1.12}
  \small
  \resizebox{\linewidth}{!}{%
  \begin{tabular}{@{}ll*{9}{r}@{}}
    \toprule
    & & \multicolumn{9}{c}{Invalid rate} \\
    \cmidrule(lr){3-11}
    Task & Stage & \multicolumn{1}{c}{\shortstack{GPT-6\\Astra}} & \multicolumn{1}{c}{\shortstack{Claude\\Opus 5}} & \multicolumn{1}{c}{\shortstack{Gemini\\3.8 Flash}} & \multicolumn{1}{c}{\shortstack{Kimi\\K3}} & \multicolumn{1}{c}{\shortstack{Grok\\4.6}} & \multicolumn{1}{c}{\shortstack{Qwen\\3.8-Max}} & \multicolumn{1}{c}{\shortstack{GLM\\5.3-Flash}} & \multicolumn{1}{c}{\shortstack{DeepSeek\\V4.1 Flash}} & \multicolumn{1}{c}{\shortstack{Doubao\\Seed 2.1 Pro}} \\
    \midrule
    \multirow{2}{*}{\Image{}} & First & 11.4\% & 10.7\% & 17.7\% & 13.7\% & 15\% & 19.7\% & 10.7\% & 19\% & 31.4\% \\
     & Final & 0.4\% & 0\% & 1\% & 0\% & 1\% & 0.3\% & 0.7\% & 3\% & 15.3\% \\
    \midrule
    \multirow{2}{*}{\TextImage{}} & First & 2.7\% & 5.3\% & 17\% & 17.7\% & 13.9\% & 21.3\% & 6.9\% & 20.7\% & 50\% \\
     & Final & 0.6\% & 0\% & 1\% & 1\% & 0\% & 1.1\% & 1.1\% & 2.1\% & 30.5\% \\
    \bottomrule
  \end{tabular}}
\end{table}

Across models and output formats, valid predictions that are successfully
decomposed retain high geometric fidelity after reassembly:
\(\mathrm{D}_{\mathrm{dec}}\) stays in the \(10^{-5}\)--\(10^{-4}\) range
and \(\mathrm{IoUV}_{\mathrm{dec}}\) exceeds \(0.92\).
At most \(5\) cases per model and format are excluded by the fidelity gate.
These results indicate that the decomposition pipeline preserves the
original predicted geometry well on the evaluated cases.

\subsection{First and final output validity}
\label{sec:invalid}

For \Image{} and \TextImage{}, the generation protocol allows up to three
generate--execute attempts: an initial output followed by at most two
corrections using execution feedback. We call the result of the first attempt
\emph{First} and the released result after this process \emph{Final}.
An output is invalid if it cannot be parsed, executed, structurally validated
or exported. Table~\ref{tab:invalid-first-final} compares the two stages.

\Text{} uses simpler single-part targets and does not adopt the
three-attempt execution-feedback correction protocol used for \Image{}
and \TextImage{}. Its mean final invalid rate is only
0.425\%%
 across all ten general-purpose
models, both specification types and JSON/OpenSCAD.

The model columns follow descending final \TextImage{} Score in
Table~\ref{tab:main-results}. The ranking broadly agrees with final validity:
all models except Doubao~Seed~2.1~Pro reach at most \(3.0\%\) final invalid
on \Image{} and \(2.1\%\) on \TextImage{}. Doubao remains the clear outlier
after correction, at \(15.3\%\) and \(30.5\%\), respectively, consistent
with its substantially lower final scores.

\subsection{Multi-turn agentic workflow}
\label{sec:agent_loop}
Beyond execution-error correction, we test whether iterative visual feedback improves reconstruction quality.
On the fixed \(100\)-case hard subset of the \Image{}/OpenSCAD task, the single-shot row in Table~\ref{tab:image-agent-loop} reuses Gemini~3.8~Flash's primary benchmark results, including execution-error correction but no visual review.
For the multi-turn setting, the model first drafts a program from the same reference image and then uses the reference image, current code, and multi-view renders to revise its output for up to ten review turns.

Multi-turn feedback raises Geo from \(0.526\) to \(0.542\), Judge from \(0.552\) to \(0.586\), and Score from \(53.88\) to \(56.38\); Topo remains \(1.000\) at the reported precision.
These gains show that iterative visual feedback improves reconstruction quality on this subset.
The model averages \(8.82\) review turns, with \(66\%\) of runs reaching the ten-turn limit, indicating frequent use of the full review budget.

\begin{table}[!t]
  \centering
  \caption{\textbf{Multi-turn agentic workflow vs.\ single-shot generation on
  the \(100\)-case hard subset of \Image{}/OpenSCAD.}
  \emph{Score} aggregates Geo and Judge;
  Topo is reported as an independent diagnostic. \emph{Turns} is the mean number of review
  turns after the initial draft (\(10\) is the maximum); ``\nodash'' marks
  the single-shot setting, for which the turn count does not apply.
  Geo, Judge and Score use bold for the better setting.}
  \label{tab:image-agent-loop}
  \setlength{\tabcolsep}{6pt}
  \renewcommand{\arraystretch}{1.1}
  \small
  \begin{tabular}{llccccc}
    \toprule
    Setting & Model & Geo & Topo & Judge & Score & Turns \\
    \midrule
    Single-shot & Gemini~3.8~Flash & 0.526 & 1.000 & 0.552 & 53.876 & \nodash \\
    Multi-turn & Gemini~3.8~Flash & \best{0.542} & 1.000 & \best{0.586} & \best{56.380} & 8.82 \\
    \bottomrule
  \end{tabular}
\end{table}

\FloatBarrier

\section{Full submetric results}
\label{app:aggregate_tables}

Tables~\ref{tab:appendix-textimage-submetrics}, \ref{tab:text-full-metrics}
and~\ref{tab:appendix-image-submetrics} report per-format metrics for
\TextImage{}, \Text{} (both specification styles) and \Image{}, respectively,
following Appendix~\ref{app:evaluation_details}.
\emph{Valid} and NoOE are percentages; other submetrics retain their native
or normalized scales. In particular, the visual-Judge submetrics report
\(1\)--\(10\) ratings \(s\) normalized to \([0,1]\) by \((s-1)/9\).
The cross-format mean of \emph{Valid} appears in Table~\ref{tab:main-results}.

\begin{table}[!hb]
  \centering
  \caption{\textbf{\TextImage{} full submetric results on the 100-case
  evaluation subset.} Panels (a) and (b) report CadQuery and OpenSCAD
  outputs, respectively. All panels include executable validity \emph{Valid}.
  \(F_1\), \(P\) and \(R\) denote \(\mathrm{PartMatchF1}\),
  \(\mathrm{PartMatchP}\) and \(\mathrm{PartMatchR}\).
  The Part bucket averages only \(F_1\) and \(\mathrm{PartFS}\);
  \(P\) and \(R\) are diagnostics.
  Normalization and aggregation follow Appendix~\ref{app:evaluation_details};
  emphasis on \emph{Geo}, \emph{Judge} and \emph{Part}-related metrics follows Table~\ref{tab:main-results}.}
  \label{tab:appendix-textimage-submetrics}

  \begin{subtable}[t]{\linewidth}
    \centering
    \caption{CadQuery output.}
    \label{tab:appendix-textimage-cadquery}
    \setlength{\tabcolsep}{1pt}
    \renewcommand{\arraystretch}{1.20}
    \footnotesize
    \resizebox{\linewidth}{!}{%
    \begin{tabular}{@{}lccccccccccccccc@{}}
      \toprule
      & \multicolumn{4}{c}{\emph{Geo}} & \multicolumn{3}{c}{\emph{Topo}}
      & \multicolumn{3}{c}{\emph{Judge}} & \multicolumn{2}{c}{\emph{PartMatch}} & \multicolumn{2}{c}{\emph{Part}} & \\
      \cmidrule(lr){2-5}\cmidrule(lr){6-8}\cmidrule(lr){9-11}\cmidrule(lr){12-13}\cmidrule(lr){14-15}
      Model & CD$\downarrow$ & \IoUV{}$\uparrow$ & F@.01$\uparrow$ & NC$\uparrow$
      & NoOE$\uparrow$ & InvN$\downarrow$ & NM$\downarrow$
      & J-Geo$\uparrow$ & J-Aes$\uparrow$ & J-Sem$\uparrow$
      & \(P\uparrow\) & \(R\uparrow\) & \(F_1\uparrow\) & PartFS\(\uparrow\) & \emph{Valid}$\uparrow$ \\
      \midrule
      GPT-6~Astra & \best{0.0009} & \best{0.632} & \best{0.478} & \best{0.727} & 77\% & 0.011 & 0.012 & \best{0.536} & \best{0.598} & \best{0.640} & \best{0.599} & \best{0.539} & \best{0.541} & \best{0.738} & 98.9\% \\
      Claude~Opus~5 & \second{0.0017} & 0.427 & \second{0.325} & \second{0.657} & 91\% & 0.000 & 0.001 & 0.416 & \second{0.526} & \second{0.636} & \second{0.502} & \second{0.426} & \second{0.444} & \second{0.681} & 100\% \\
      Gemini~3.8~Flash & 0.0020 & \second{0.429} & 0.310 & 0.637 & 89\% & 0.021 & 0.022 & \second{0.431} & 0.516 & 0.628 & 0.460 & 0.366 & 0.391 & 0.616 & 97.9\% \\
      Kimi~K3 & 0.0025 & 0.359 & 0.271 & 0.618 & 74\% & 0.021 & 0.022 & 0.366 & 0.454 & 0.608 & 0.415 & 0.423 & 0.409 & 0.619 & 97.9\% \\
      Grok~4.6 & 0.0023 & 0.381 & 0.290 & 0.635 & 86\% & 0.000 & 0.002 & 0.367 & 0.474 & 0.584 & 0.407 & 0.322 & 0.338 & 0.609 & 100\% \\
      Qwen3.8-Max & 0.0029 & 0.337 & 0.273 & 0.612 & 80\% & 0.022 & 0.022 & 0.349 & 0.437 & 0.576 & 0.392 & 0.364 & 0.368 & 0.576 & 97.8\% \\
      GLM-5.3-Flash & 0.0023 & 0.374 & 0.276 & 0.610 & 82\% & 0.023 & 0.024 & 0.347 & 0.462 & 0.587 & 0.362 & 0.335 & 0.342 & 0.566 & 97.7\% \\
      DeepSeek~V4.1~Flash & 0.0029 & 0.383 & 0.269 & 0.594 & 81\% & 0.032 & 0.037 & 0.344 & 0.449 & 0.569 & 0.381 & 0.330 & 0.340 & 0.591 & 96.8\% \\
      Doubao~Seed~2.1~Pro & 0.0072 & 0.119 & 0.102 & 0.240 & 37\% & 0.596 & 0.598 & 0.128 & 0.156 & 0.208 & 0.159 & 0.118 & 0.130 & 0.217 & 41\% \\
      \bottomrule
    \end{tabular}}
  \end{subtable}

  \vspace{1.2em}

  \begin{subtable}[t]{\linewidth}
    \centering
    \caption{OpenSCAD output.}
    \label{tab:appendix-textimage-openscad}
    \setlength{\tabcolsep}{1pt}
    \renewcommand{\arraystretch}{1.20}
    \footnotesize
    \resizebox{\linewidth}{!}{%
    \begin{tabular}{@{}lccccccccccccccc@{}}
      \toprule
      & \multicolumn{4}{c}{\emph{Geo}} & \multicolumn{3}{c}{\emph{Topo}}
      & \multicolumn{3}{c}{\emph{Judge}} & \multicolumn{2}{c}{\emph{PartMatch}} & \multicolumn{2}{c}{\emph{Part}} & \\
      \cmidrule(lr){2-5}\cmidrule(lr){6-8}\cmidrule(lr){9-11}\cmidrule(lr){12-13}\cmidrule(lr){14-15}
      Model & CD$\downarrow$ & \IoUV{}$\uparrow$ & F@.01$\uparrow$ & NC$\uparrow$
      & NoOE$\uparrow$ & InvN$\downarrow$ & NM$\downarrow$
      & J-Geo$\uparrow$ & J-Aes$\uparrow$ & J-Sem$\uparrow$
      & \(P\uparrow\) & \(R\uparrow\) & \(F_1\uparrow\) & PartFS\(\uparrow\) & \emph{Valid}$\uparrow$ \\
      \midrule
      GPT-6~Astra & \best{0.0007} & \best{0.583} & \best{0.448} & \best{0.723} & 100\% & 0.000 & 0.000 & \best{0.539} & \best{0.614} & \best{0.666} & \best{0.573} & \best{0.489} & \best{0.501} & \best{0.710} & 100\% \\
      Claude~Opus~5 & 0.0019 & 0.404 & 0.317 & 0.647 & 100\% & 0.000 & 0.000 & 0.392 & 0.517 & 0.637 & \second{0.461} & 0.392 & 0.414 & 0.631 & 100\% \\
      Gemini~3.8~Flash & 0.0018 & \second{0.437} & \second{0.329} & \second{0.659} & 100\% & 0.000 & 0.001 & \second{0.424} & \second{0.532} & \second{0.646} & 0.455 & 0.357 & 0.380 & \second{0.632} & 100\% \\
      Kimi~K3 & 0.0021 & 0.379 & 0.287 & 0.642 & 100\% & 0.000 & 0.001 & 0.369 & 0.471 & 0.628 & 0.436 & 0.403 & 0.406 & 0.615 & 100\% \\
      Grok~4.6 & \second{0.0017} & 0.408 & 0.322 & 0.656 & 100\% & 0.000 & 0.001 & 0.398 & 0.513 & 0.638 & 0.397 & 0.325 & 0.338 & 0.616 & 100\% \\
      Qwen3.8-Max & 0.0019 & 0.401 & 0.297 & 0.636 & 100\% & 0.000 & 0.000 & 0.371 & 0.481 & 0.623 & 0.447 & \second{0.409} & \second{0.416} & 0.625 & 100\% \\
      GLM-5.3-Flash & 0.0022 & 0.379 & 0.287 & 0.635 & 100\% & 0.000 & 0.000 & 0.370 & 0.498 & 0.621 & 0.373 & 0.330 & 0.341 & 0.587 & 100\% \\
      DeepSeek~V4.1~Flash & 0.0023 & 0.355 & 0.266 & 0.617 & 99\% & 0.010 & 0.011 & 0.350 & 0.472 & 0.596 & 0.356 & 0.301 & 0.315 & 0.577 & 99\% \\
      Doubao~Seed~2.1~Pro & 0.0033 & 0.308 & 0.236 & 0.595 & 98\% & 0.020 & 0.021 & 0.287 & 0.357 & 0.494 & 0.337 & 0.262 & 0.279 & 0.541 & 98\% \\
      \bottomrule
    \end{tabular}}
  \end{subtable}

\end{table}

\clearpage

\begin{table}[!p]
  \centering
  \caption{\textbf{\Text{} full submetric results on the 100-case
  evaluation subset.} Panel (a) reports descriptive specifications for
  JSON and OpenSCAD; panels (b) and (c) report parametric specifications
  for JSON and OpenSCAD, respectively. All panels include executable
  validity \emph{Valid}. The domain-specific baseline Text2CAD uses
  native JSON and is listed separately below the general-purpose models.
  ``\nodash'' denotes an unevaluated model--format pair.
  Normalization and aggregation follow Appendix~\ref{app:evaluation_details};
  emphasis on \emph{Geo} and \emph{Judge} metrics follows Table~\ref{tab:main-results}.}
  \label{tab:text-full-metrics}

  \vspace{0.15em}

  \begin{subtable}[t]{\linewidth}
    \centering
    \caption{\TextDesc{}.}
    \label{tab:text-full-desc-metrics}
    \setlength{\tabcolsep}{5pt}
    \renewcommand{\arraystretch}{1.20}
    \footnotesize
    \begin{tabular*}{\linewidth}{@{\extracolsep{\fill}}lcccccc@{}}
      \toprule
      & \multicolumn{3}{c}{\emph{JSON}} & \multicolumn{3}{c}{\emph{OpenSCAD}} \\
      \cmidrule(lr){2-4}\cmidrule(lr){5-7}
      Model & QA-S$\uparrow$ & J-Sem$\uparrow$ & \emph{Valid}$\uparrow$ & QA-S$\uparrow$ & J-Sem$\uparrow$ & \emph{Valid}$\uparrow$ \\
      \midrule
      GPT-6~Astra & \best{0.978} & \best{0.877} & 100\% & \second{0.993} & \best{0.859} & 100\% \\
      Claude~Opus~5 & 0.932 & 0.726 & 100\% & 0.975 & 0.751 & 100\% \\
      Gemini~3.8~Flash & 0.960 & 0.751 & 100\% & \best{0.995} & 0.758 & 100\% \\
      Kimi~K3 & 0.897 & 0.764 & 97\% & 0.990 & \second{0.826} & 100\% \\
      Grok~4.6 & 0.960 & 0.726 & 100\% & \best{0.995} & 0.756 & 100\% \\
      Qwen3.8-Max & \second{0.963} & 0.772 & 100\% & 0.988 & 0.822 & 100\% \\
      GLM-5.3 & 0.900 & \second{0.791} & 100\% & 0.935 & 0.783 & 99\% \\
      GLM-5.3-Flash & 0.905 & 0.757 & 98\% & 0.870 & 0.712 & 97\% \\
      DeepSeek~V4.1~Flash & 0.910 & 0.771 & 100\% & 0.975 & 0.793 & 100\% \\
      Doubao~Seed~2.1~Pro & 0.880 & 0.771 & 99\% & 0.938 & 0.780 & 100\% \\
      \midrule
      Text2CAD & 0.058 & 0.200 & 91\% & \nodash & \nodash & \nodash \\
      \bottomrule
    \end{tabular*}
  \end{subtable}

  \vspace{0.3em}

  \begin{subtable}[t]{\linewidth}
    \centering
    \caption{\TextParam{}, JSON.}
    \label{tab:text-full-param-json}
    \setlength{\tabcolsep}{5pt}
    \renewcommand{\arraystretch}{1.20}
    \footnotesize
    \resizebox{\linewidth}{!}{%
    \begin{tabular}{@{}lcccccccccc@{}}
      \toprule
      & \multicolumn{4}{c}{\emph{Geo}} & \multicolumn{3}{c}{\emph{Topo}} & \multicolumn{2}{c}{\emph{Judge}} & \\
      \cmidrule(lr){2-5}\cmidrule(lr){6-8}\cmidrule(lr){9-10}
      Model
        & CD$\downarrow$ & \(\IoUC\uparrow\)
        & F@.01$\uparrow$ & NC$\uparrow$
        & NoOE$\uparrow$ & InvN$\downarrow$ & NM$\downarrow$
        & QA-S$\uparrow$ & QA-P$\uparrow$ & \emph{Valid}$\uparrow$ \\
      \midrule
      GPT-6~Astra & 0.0013 & 0.618 & 0.412 & 0.756 & 96\% & 0.0000 & 0.0004 & \best{0.917} & \best{0.907} & 100\% \\
      Claude~Opus~5 & \second{0.0012} & 0.603 & 0.402 & 0.746 & 97\% & 0.0000 & 0.0005 & 0.868 & 0.895 & 100\% \\
      Gemini~3.8~Flash & \best{0.0011} & \second{0.620} & \second{0.428} & \second{0.767} & 96\% & 0.0000 & 0.0008 & 0.858 & 0.892 & 100\% \\
      Kimi~K3 & 0.0013 & 0.602 & 0.399 & 0.742 & 95\% & 0.0100 & 0.0117 & 0.868 & 0.894 & 99\% \\
      Grok~4.6 & \best{0.0011} & \best{0.623} & \best{0.436} & \best{0.768} & 97\% & 0.0000 & 0.0009 & 0.853 & 0.890 & 100\% \\
      Qwen3.8-Max & \best{0.0011} & 0.610 & 0.409 & 0.762 & 94\% & 0.0000 & 0.0016 & \second{0.877} & \second{0.905} & 100\% \\
      GLM-5.3 & 0.0013 & 0.574 & 0.382 & 0.733 & 97\% & 0.0000 & 0.0012 & 0.860 & 0.875 & 100\% \\
      GLM-5.3-Flash & 0.0016 & 0.518 & 0.354 & 0.711 & 93\% & 0.0200 & 0.0211 & 0.843 & 0.865 & 98\% \\
      DeepSeek~V4.1~Flash & 0.0014 & 0.588 & 0.398 & 0.742 & 95\% & 0.0100 & 0.0118 & 0.820 & 0.886 & 99\% \\
      Doubao~Seed~2.1~Pro & 0.0015 & 0.572 & 0.389 & 0.738 & 95\% & 0.0100 & 0.0111 & 0.848 & 0.871 & 99\% \\
      \midrule
      Text2CAD & 0.0056 & 0.173 & 0.081 & 0.473 & 98\% & 0.0200 & 0.0278 & 0.077 & 0.163 & 98\% \\
      \bottomrule
    \end{tabular}}
  \end{subtable}

  \vspace{0.3em}

  \begin{subtable}[t]{\linewidth}
    \centering
    \caption{\TextParam{}, OpenSCAD.}
    \label{tab:text-full-param-openscad}
    \setlength{\tabcolsep}{5pt}
    \renewcommand{\arraystretch}{1.20}
    \footnotesize
    \resizebox{\linewidth}{!}{%
    \begin{tabular}{@{}lcccccccccc@{}}
      \toprule
      & \multicolumn{4}{c}{\emph{Geo}} & \multicolumn{3}{c}{\emph{Topo}} & \multicolumn{2}{c}{\emph{Judge}} & \\
      \cmidrule(lr){2-5}\cmidrule(lr){6-8}\cmidrule(lr){9-10}
      Model
        & CD$\downarrow$ & \(\IoUC\uparrow\)
        & F@.01$\uparrow$ & NC$\uparrow$
        & NoOE$\uparrow$ & InvN$\downarrow$ & NM$\downarrow$
        & QA-S$\uparrow$ & QA-P$\uparrow$ & \emph{Valid}$\uparrow$ \\
      \midrule
      GPT-6~Astra & \best{0.0007} & \best{0.676} & \best{0.445} & \best{0.790} & 100\% & 0.0000 & 0.0000 & \best{0.945} & \second{0.944} & 100\% \\
      Claude~Opus~5 & \second{0.0009} & 0.653 & 0.432 & 0.775 & 100\% & 0.0000 & 0.0008 & \best{0.945} & 0.934 & 100\% \\
      Gemini~3.8~Flash & \best{0.0007} & \second{0.671} & \second{0.437} & \second{0.786} & 100\% & 0.0000 & 0.0001 & 0.917 & 0.936 & 100\% \\
      Kimi~K3 & \second{0.0009} & 0.647 & 0.418 & 0.764 & 100\% & 0.0000 & 0.0005 & 0.930 & \best{0.945} & 100\% \\
      Grok~4.6 & \second{0.0009} & 0.652 & 0.430 & 0.780 & 100\% & 0.0000 & 0.0002 & 0.885 & 0.929 & 100\% \\
      Qwen3.8-Max & \second{0.0009} & 0.650 & 0.416 & 0.773 & 100\% & 0.0000 & 0.0000 & 0.910 & 0.921 & 100\% \\
      GLM-5.3 & 0.0010 & 0.622 & 0.398 & 0.756 & 100\% & 0.0000 & 0.0007 & \second{0.932} & 0.929 & 100\% \\
      GLM-5.3-Flash & 0.0017 & 0.543 & 0.336 & 0.707 & 99\% & 0.0100 & 0.0105 & 0.848 & 0.906 & 99\% \\
      DeepSeek~V4.1~Flash & 0.0012 & 0.610 & 0.401 & 0.748 & 99\% & 0.0100 & 0.0114 & 0.887 & 0.919 & 99\% \\
      Doubao~Seed~2.1~Pro & 0.0012 & 0.593 & 0.388 & 0.747 & 100\% & 0.0000 & 0.0003 & 0.885 & 0.922 & 100\% \\
      \bottomrule
    \end{tabular}}
  \end{subtable}

\end{table}

\begin{table}[!p]
  \centering
  \caption{\textbf{\Image{} full submetric results on the 100-case
  evaluation subset.} Panels (a), (b) and (c) report CadQuery, OpenSCAD
  and Three.js outputs, respectively. All panels include executable
  validity \emph{Valid}. The domain-specific baselines \textsc{Cadrille} and
  \textsc{CAD-Coder} use native CadQuery, are listed separately below
  the general-purpose models.
  Normalization and aggregation follow Appendix~\ref{app:evaluation_details};
  emphasis on \emph{Geo} and \emph{Judge} metrics follows Table~\ref{tab:main-results}.}
  \label{tab:appendix-image-submetrics}

  \vspace{0.35em}

  \begin{subtable}[t]{\linewidth}
    \centering
    \caption{CadQuery output.}
    \label{tab:appendix-image-cadquery}
    \setlength{\tabcolsep}{4pt}
    \renewcommand{\arraystretch}{1.20}
    \footnotesize
    \resizebox{\linewidth}{!}{%
    \begin{tabular}{@{}lccccccccccc@{}}
      \toprule
      & \multicolumn{4}{c}{\emph{Geo}} & \multicolumn{3}{c}{\emph{Topo}} & \multicolumn{3}{c}{\emph{Judge}} & \\
      \cmidrule(lr){2-5}\cmidrule(lr){6-8}\cmidrule(lr){9-11}
      Model & CD$\downarrow$ & \(\IoUV\uparrow\) & F@.01$\uparrow$ & NC$\uparrow$
        & NoOE$\uparrow$ & InvN$\downarrow$ & NM$\downarrow$
        & J-Geo$\uparrow$ & J-Aes$\uparrow$ & J-Sem$\uparrow$ & \emph{Valid}$\uparrow$ \\
      \midrule
      GPT-6~Astra & \best{0.0012} & \best{0.414} & \best{0.315} & \best{0.599} & 81\% & 0.012 & 0.015 & \best{0.508} & \best{0.628} & \best{0.661} & 98.8\% \\
      Gemini~3.8~Flash & \second{0.0017} & \second{0.351} & \second{0.257} & \second{0.566} & 93\% & 0.021 & 0.022 & \second{0.436} & \second{0.531} & \second{0.646} & 98\% \\
      Claude~Opus~5 & 0.0021 & 0.316 & 0.240 & 0.561 & 88\% & 0.000 & 0.002 & 0.394 & 0.518 & 0.628 & 100\% \\
      Kimi~K3 & 0.0021 & 0.307 & 0.223 & 0.551 & 95\% & 0.000 & 0.005 & 0.382 & 0.508 & 0.632 & 100\% \\
      Grok~4.6 & 0.0027 & 0.274 & 0.216 & 0.525 & 90\% & 0.031 & 0.031 & 0.364 & 0.462 & 0.598 & 97\% \\
      Qwen3.8-Max & 0.0021 & 0.312 & 0.221 & 0.541 & 92\% & 0.010 & 0.013 & 0.387 & 0.481 & 0.623 & 99\% \\
      GLM-5.3-Flash & 0.0021 & 0.301 & 0.216 & 0.553 & 93\% & 0.000 & 0.000 & 0.361 & 0.480 & 0.616 & 100\% \\
      DeepSeek~V4.1~Flash & 0.0033 & 0.248 & 0.184 & 0.484 & 86\% & 0.092 & 0.092 & 0.310 & 0.403 & 0.514 & 91\% \\
      Doubao~Seed~2.1~Pro & 0.0061 & 0.116 & 0.101 & 0.287 & 46\% & 0.450 & 0.451 & 0.187 & 0.227 & 0.318 & 55\% \\
      \midrule
      \textsc{Cadrille} & 0.0074 & 0.056 & 0.076 & 0.372 & 71\% & 0.237 & 0.238 & 0.167 & 0.163 & 0.167 & 77\% \\
      \textsc{CAD-Coder} & 0.0091 & 0.018 & 0.024 & 0.107 & 20\% & 0.793 & 0.794 & 0.044 & 0.041 & 0.043 & 20.7\% \\
      \bottomrule
    \end{tabular}}
  \end{subtable}

  \vspace{0.25em}

  \begin{subtable}[t]{\linewidth}
    \centering
    \caption{OpenSCAD output.}
    \label{tab:appendix-image-openscad}
    \setlength{\tabcolsep}{4pt}
    \renewcommand{\arraystretch}{1.20}
    \footnotesize
    \resizebox{\linewidth}{!}{%
    \begin{tabular}{@{}lccccccccccc@{}}
      \toprule
      & \multicolumn{4}{c}{\emph{Geo}} & \multicolumn{3}{c}{\emph{Topo}} & \multicolumn{3}{c}{\emph{Judge}} & \\
      \cmidrule(lr){2-5}\cmidrule(lr){6-8}\cmidrule(lr){9-11}
      Model & CD$\downarrow$ & \(\IoUV\uparrow\) & F@.01$\uparrow$ & NC$\uparrow$
        & NoOE$\uparrow$ & InvN$\downarrow$ & NM$\downarrow$
        & J-Geo$\uparrow$ & J-Aes$\uparrow$ & J-Sem$\uparrow$ & \emph{Valid}$\uparrow$ \\
      \midrule
      GPT-6~Astra & \best{0.0011} & \best{0.401} & \best{0.283} & \best{0.601} & 100\% & 0.000 & 0.000 & \best{0.510} & \best{0.633} & \best{0.677} & 100\% \\
      Gemini~3.8~Flash & \second{0.0018} & \second{0.358} & \second{0.258} & \second{0.566} & 100\% & 0.000 & 0.001 & \second{0.436} & \second{0.556} & \second{0.664} & 100\% \\
      Claude~Opus~5 & \second{0.0018} & 0.327 & 0.234 & 0.557 & 100\% & 0.000 & 0.000 & 0.404 & 0.551 & 0.648 & 100\% \\
      Kimi~K3 & 0.0020 & 0.315 & 0.226 & 0.557 & 100\% & 0.000 & 0.000 & 0.399 & 0.532 & 0.643 & 100\% \\
      Grok~4.6 & 0.0020 & 0.312 & 0.232 & 0.552 & 100\% & 0.000 & 0.000 & 0.400 & 0.523 & 0.641 & 100\% \\
      Qwen3.8-Max & 0.0021 & 0.312 & 0.214 & 0.549 & 100\% & 0.000 & 0.000 & 0.384 & 0.481 & 0.637 & 100\% \\
      GLM-5.3-Flash & 0.0022 & 0.291 & 0.203 & 0.529 & 98\% & 0.021 & 0.021 & 0.359 & 0.493 & 0.612 & 98\% \\
      DeepSeek~V4.1~Flash & 0.0023 & 0.300 & 0.214 & 0.548 & 100\% & 0.000 & 0.000 & 0.366 & 0.516 & 0.624 & 100\% \\
      Doubao~Seed~2.1~Pro & 0.0027 & 0.293 & 0.192 & 0.536 & 99\% & 0.010 & 0.010 & 0.321 & 0.417 & 0.579 & 99\% \\
      \bottomrule
    \end{tabular}}
  \end{subtable}

  \vspace{0.25em}

  \begin{subtable}[t]{\linewidth}
    \centering
    \caption{Three.js output.}
    \label{tab:appendix-image-threejs}
    \setlength{\tabcolsep}{4pt}
    \renewcommand{\arraystretch}{1.20}
    \footnotesize
    \resizebox{\linewidth}{!}{%
    \begin{tabular}{@{}lccccccccccc@{}}
      \toprule
      & \multicolumn{4}{c}{\emph{Geo}} & \multicolumn{3}{c}{\emph{Topo}} & \multicolumn{3}{c}{\emph{Judge}} & \\
      \cmidrule(lr){2-5}\cmidrule(lr){6-8}\cmidrule(lr){9-11}
      Model & CD$\downarrow$ & \(\IoUV\uparrow\) & F@.01$\uparrow$ & NC$\uparrow$
        & NoOE$\uparrow$ & InvN$\downarrow$ & NM$\downarrow$
        & J-Geo$\uparrow$ & J-Aes$\uparrow$ & J-Sem$\uparrow$ & \emph{Valid}$\uparrow$ \\
      \midrule
      GPT-6~Astra & \best{0.0007} & \best{0.465} & \best{0.343} & \best{0.619} & 52\% & 0.002 & 0.006 & \best{0.548} & \best{0.643} & \best{0.676} & 100\% \\
      Gemini~3.8~Flash & 0.0018 & 0.270 & \second{0.258} & 0.555 & 16\% & 0.012 & 0.016 & \second{0.432} & 0.543 & 0.647 & 99\% \\
      Claude~Opus~5 & \second{0.0015} & \second{0.378} & 0.246 & \second{0.563} & 44\% & 0.001 & 0.003 & 0.422 & \second{0.559} & \second{0.656} & 100\% \\
      Kimi~K3 & 0.0020 & 0.331 & 0.240 & 0.552 & 40\% & 0.000 & 0.012 & 0.380 & 0.523 & 0.636 & 100\% \\
      Grok~4.6 & 0.0019 & 0.346 & 0.248 & 0.555 & 31\% & 0.001 & 0.002 & 0.396 & 0.528 & 0.640 & 100\% \\
      Qwen3.8-Max & 0.0018 & 0.323 & 0.236 & 0.554 & 61\% & 0.001 & 0.008 & 0.390 & 0.491 & 0.644 & 100\% \\
      GLM-5.3-Flash & 0.0020 & 0.314 & 0.229 & 0.553 & 43\% & 0.001 & 0.007 & 0.396 & 0.520 & 0.651 & 100\% \\
      DeepSeek~V4.1~Flash & 0.0021 & 0.332 & 0.231 & 0.545 & 54\% & 0.002 & 0.006 & 0.362 & 0.487 & 0.612 & 100\% \\
      Doubao~Seed~2.1~Pro & 0.0021 & 0.324 & 0.223 & 0.548 & 36\% & 0.002 & 0.007 & 0.357 & 0.439 & 0.614 & 100\% \\
      \bottomrule
    \end{tabular}}
  \end{subtable}

\end{table}

\clearpage

\begin{figure}[!p]
\section{Qualitative visualizations}
\label{app:qualitative}
  \centering
  \begin{subfigure}[t]{\linewidth}
    \centering
    \includegraphics[width=\linewidth,height=0.80\textheight,keepaspectratio]{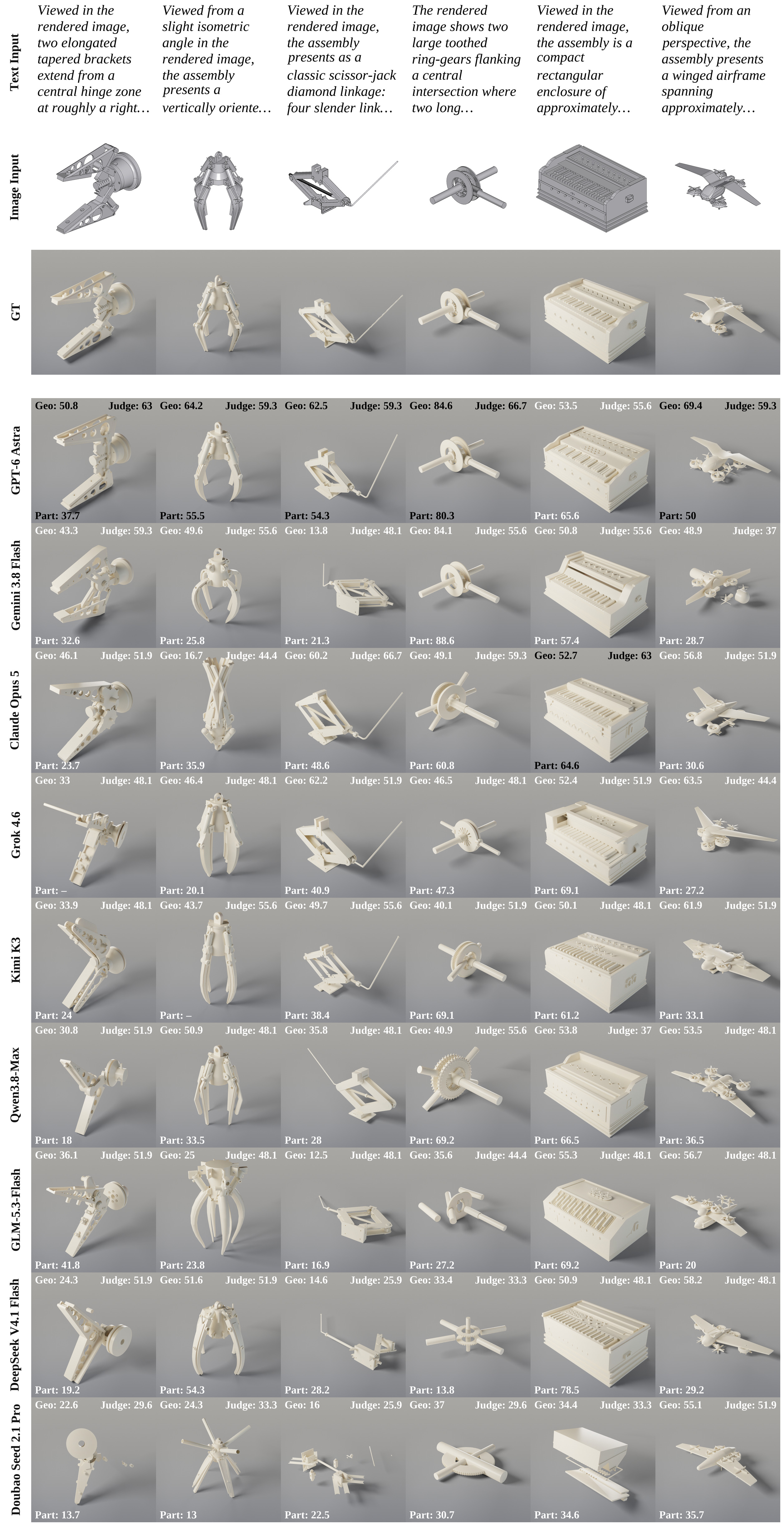}
    \caption{\TextImage{}: each column pairs the assembly input, a
    render together with the part and assembly text, with the
    reconstructions of the selected models, all in
    OpenSCAD.}
    \label{fig:qual-assembly}
  \end{subfigure}
  \caption{\textbf{Qualitative reconstructions on the three \sysname{} tasks.}
  Columns are cases; rows are the task input, the ground truth and
  the selected models.  The general-purpose models
  produce OpenSCAD, while the domain-specific baselines are shown in
  their native formats.  Each output is annotated with the per-case
  bucket scores; ``invalid'' marks a prediction that
  failed to execute and ``Part:~\nodash'' one that failed the
  \emph{Part} fidelity gate.  Sub-figures (a)--(c) cover \TextImage{},
  \Text{} and \Image{}, respectively.}
  \label{fig:qual-all}
\end{figure}

\begin{figure}[!p]
  \ContinuedFloat
  \centering
  \begin{subfigure}[t]{\linewidth}
    \centering
    \includegraphics[width=\linewidth,height=0.88\textheight,keepaspectratio]{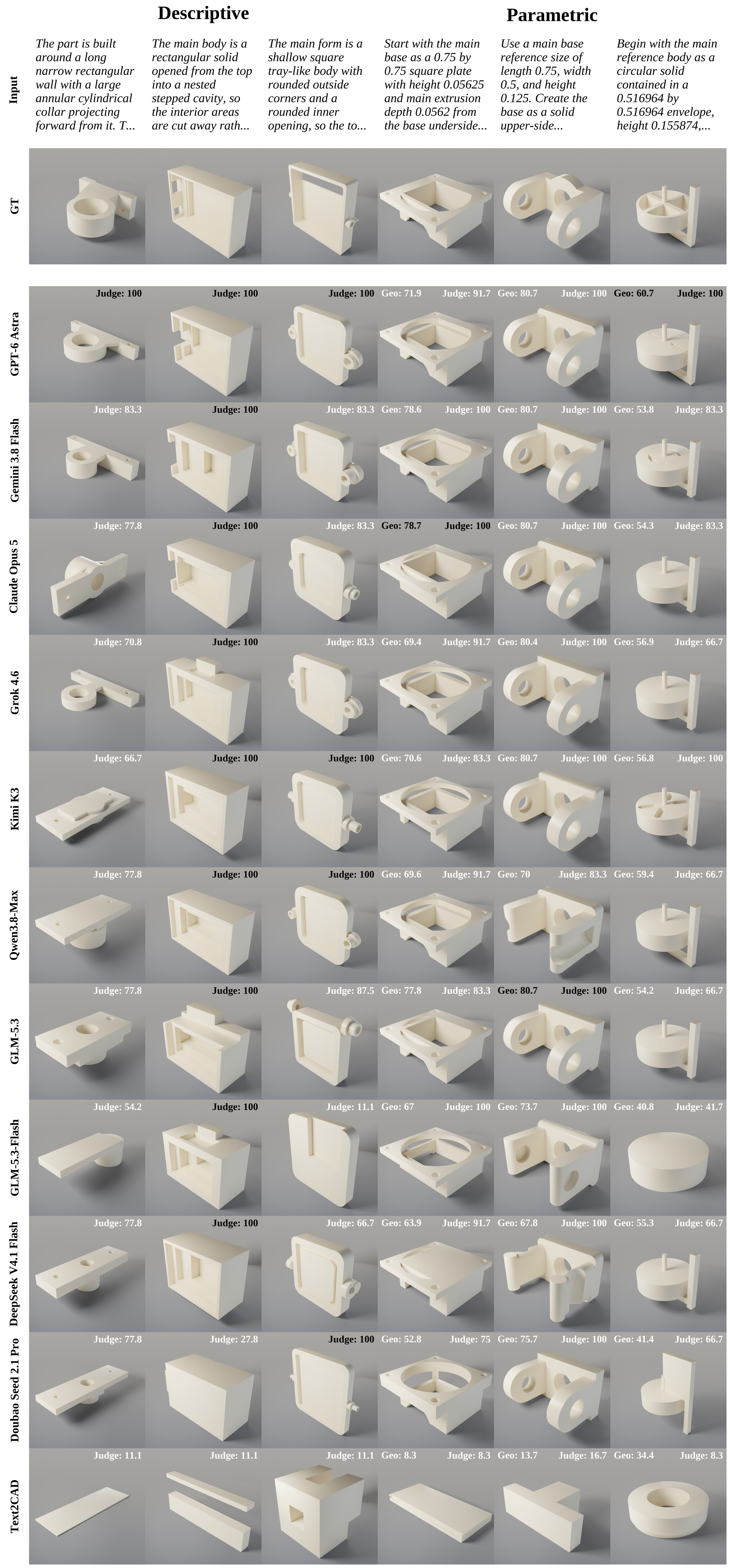}
    \caption{\Text{}: three \TextDescShort{} and three
    \TextParamShort{} cases, where ``Desc.''\ and ``Param.''\ denote the
    descriptive and parametric setting, and the domain-specific
    baseline is Text2CAD.}
    \label{fig:qual-text}
  \end{subfigure}
  \caption{(Continued.)}
\end{figure}

\begin{figure}[!p]
  \ContinuedFloat
  \centering
  \begin{subfigure}[t]{\linewidth}
    \centering
    \includegraphics[width=\linewidth,height=0.88\textheight,keepaspectratio]{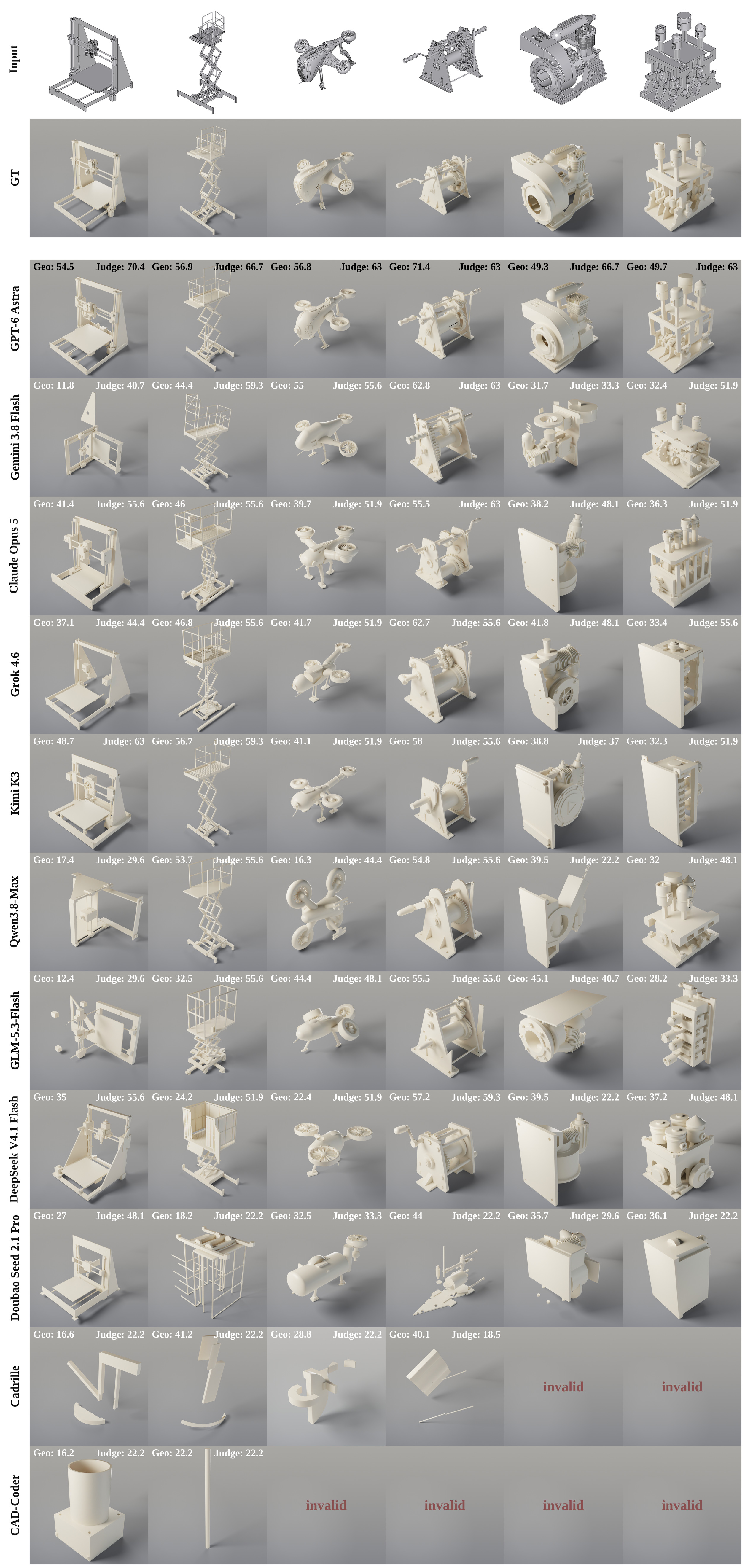}
    \caption{\Image{}: each column pairs an input render with the
    reconstructions of the selected models; the
    domain-specific baselines are \textsc{Cadrille} and CAD-Coder.}
    \label{fig:qual-image}
  \end{subfigure}
  \caption{(Continued.)}
\end{figure}

\end{document}